\documentclass[10pt,twocolumn,letterpaper]{article}
\usepackage{titling}        
\usepackage{iccv}
\usepackage{times}
\usepackage{epsfig}
\usepackage{graphicx}
\usepackage{amsmath}
\usepackage{amssymb}

\usepackage{url}            
\usepackage{booktabs}       
\usepackage{amsfonts}       
\usepackage{nicefrac}       
\usepackage{microtype}      
\usepackage{multirow}
\usepackage{makecell}
\usepackage{array}
\usepackage{wrapfig}
\usepackage[dvipsnames]{xcolor}
\usepackage{colortbl}
\usepackage{graphicx, caption, subcaption}

\usepackage[pagebackref=true,breaklinks=true,letterpaper=true,colorlinks,bookmarks=false]{hyperref}
\DeclareMathSizes{10}{9}{7}{5}

\iccvfinalcopy 

\begin{document}
\title{BossNAS: Exploring Hybrid CNN-transformers with Block-wisely Self-supervised Neural Architecture Search}

\author{Changlin Li$^{1}$, \quad Tao Tang$^2$, \quad Guangrun Wang$^{3,4}$, \quad Jiefeng Peng$^{3}$, \quad Bing Wang$^{5}$,\\ \quad Xiaodan Liang$^{2}$\thanks{Corresponding Author.}, \quad Xiaojun Chang$^{6}$\\
\small$^1$GORSE Lab, Dept. of DSAI,
Monash University \quad \small$^2$Sun Yat-sen University\quad \small$^3$DarkMatter AI Research \\ \quad \small$^4$University of Oxford \quad \small$^5$Alibaba Group \quad \small $^6$RMIT University\\
{\tt\small changlin.li@monash.edu, }\\{\tt\small \{trent.tangtao,wanggrun,jiefengpeng,xdliang328\}@gmail.com,}\\{\tt\small fengquan.wb@alibaba-inc.com, xiaojun.chang@rmit.edu.au}
\vspace{-10pt}}

\maketitle

\begin{abstract}\vspace{-5pt}
A myriad of recent breakthroughs in hand-crafted neural architectures for visual recognition have highlighted the urgent need to explore hybrid architectures consisting of diversified building blocks. Meanwhile, neural architecture search methods are surging with an expectation to reduce human efforts. However, whether NAS methods can efficiently and effectively handle diversified search spaces with disparate candidates (\textit{e.g.} CNNs and transformers) is still an open question. In this work, we present \textbf{B}l\textbf{o}ck-wisely \textbf{S}elf-\textbf{s}upervised \textbf{N}eural \textbf{A}rchitecture \textbf{S}earch (BossNAS), an unsupervised NAS method that addresses the problem of inaccurate architecture rating caused by large weight-sharing space and biased supervision in previous methods. More specifically, we factorize the search space into blocks and utilize a novel self-supervised training scheme, named ensemble bootstrapping, to train each block separately before searching them as a whole towards the population center. Additionally, we present HyTra search space, a fabric-like hybrid CNN-transformer search space with searchable down-sampling positions.
On this challenging search space, our searched model, BossNet-T, achieves up to 82.5\% accuracy on ImageNet, surpassing EfficientNet by 2.4\% with comparable compute time. Moreover, our method achieves superior architecture rating accuracy with 0.78 and 0.76 Spearman correlation on the canonical MBConv search space with ImageNet and on NATS-Bench size search space with CIFAR-100, respectively, surpassing \textit{state-of-the-art} NAS methods.\footnote{Code: \url{https://github.com/changlin31/BossNAS}.}
\vspace{-5pt}%
\end{abstract}%
%
\begin{figure}
\vspace{-10pt}
\centering
\begin{subfigure}[t]{.9\linewidth}
    \centering\includegraphics[width=.9\linewidth]{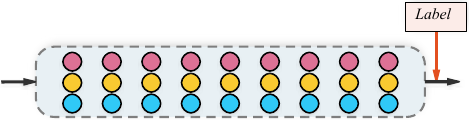}\vspace{-2pt}
    \caption{\small NAS with large weight-sharing space ($3^9$).}\label{fig:intro_a}
  \end{subfigure}\\\vspace{2pt}
  \begin{subfigure}[t]{.9\linewidth}
    \centering\includegraphics[width=.9\linewidth]{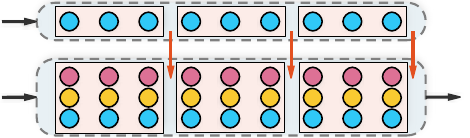}\vspace{-2pt}
    \caption{\small Block-wise NAS with biased supervision.}\label{fig:intro_b}
  \end{subfigure}\\\vspace{2pt}
  \begin{subfigure}[t]{.9\linewidth}
    \centering\includegraphics[width=.9\linewidth]{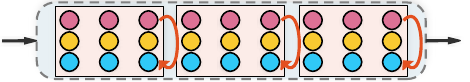}\vspace{-2pt}
    \caption{\small Block-wisely self-supervised NAS.}\label{fig:intro_c}
  \end{subfigure}
  \vspace{-5pt}
  \caption{Comparision of three NAS schemes. Red arrows represent the supervision during training and searching.}
\vspace{-15pt}
\end{figure}
\vspace{-15pt}
\section{Introduction}
\vspace{-5pt}

The development of neural network architectures has brought about significant progress in a wide range of visual recognition tasks over the past several years. Representative examples of such models include ResNet \cite{he2016resnet}, SENet \cite{Hu2019SqueezeandExcitationN}, MobileNet \cite{howard2017mobilenets} and EfficientNet \cite{Tan2019EfficientNetRM}.
Recently, the newly emerging attention-based architectures are coming to the forefront in the vision field, challenging the dominance of convolutional neural networks (CNNs). This exciting breakthrough in vision transformers led by ViT \cite{dosovitskiy2021vit} and DETR \cite{carion2020detr}, are achieving competitive performance on various vision tasks, such as image classification \cite{dosovitskiy2021vit,touvron2020deit,yuan2021t2t,chu2021conditional,Chen2021AutoFormerST}, object detection \cite{carion2020detr,zhu2021deformabledetr,srinivas2021bottleneck}, semantic segmentation \cite{zheng2020setr}, and others \cite{he2021transreid,parmar2018imagetransformer,jiang2021transgan}. As suggested by prior works \cite{dosovitskiy2021vit,srinivas2021bottleneck,bello2021lambdanetworks}, hybrids of CNNs and transformers can outperform both pure transformers and pure CNNs. 

Despite the large advances brought about by network design, manually finding well-optimized hybrid architectures can be challenging, especially as the number of design choices increases. Neural Architecture Search (NAS) is a popular approach to reducing the human effort in network architecture design by automatically searching for optimal architectures in a predefined search space. Representative success in performing NAS on manually designed building blocks include MobileNetV3 \cite{Howard2019SearchingFM}, EfficientNet \cite{Tan2019EfficientNetRM}, \textit{etc.} These works are searched by \textit{multi-trial} NAS methods \cite{Tan2018MnasNetPN,zoph2016neural, baker2016designing, zhong2018practical, chen2018reinforced, negrinho2017deeparchitect}, which are computationally prohibitive (costing thousands of GPU days). Recent \textit{weight-sharing} NAS methods \cite{brock2017smash,Pham2018EfficientNA,bender2018understanding,Liu2018DARTSDA} encode the entire search space as a weight-sharing \textit{supernet} to avoid repetitive training of candidate networks, thus largely reducing the search cost.

However, as shown in Fig. \ref{fig:intro_a}, architecture search spaces with layer-level granularity grow exponentially with increased network depth, which has been identified (in \cite{Li2020Blockwisely_cvpr,li2020improving}) as the main culprit of inaccurate architecture rating\footnote{In this work, \textit{architecture rating accuracy} refers to the correlation of the predicted architecture ranking and the ground truth architecture ranking.\vspace{-10pt}} in weight-sharing NAS methods.
To reduce the size of the large weight-sharing space, previous works \cite{Li2020Blockwisely_cvpr, moons2020donna} factorize the search space into blocks and use a pretrained teacher model to provide block-wise supervision (Fig. \ref{fig:intro_b}).
Despite their high ranking correlation and high efficiency, we find (in Sec. \ref{sec:exp}) their results to be highly correlated with the teacher architecture. As illustrated in Fig. \ref{fig:intro_b}, when training by a teacher with blue nodes, candidate architectures with more blue nodes tend to get higher ranks in these methods. This limits its application on diversified search spaces with disparate candidates, such as CNNs and transformers.

On the other hand, unsupervised NAS \cite{liu2020unnas} has recently emerged as an interesting research topic. Without access to any human-annotated labels, unsupervised NAS methods (optimized with pretext tasks \cite{liu2020unnas} or random labels \cite{zhang2021randomlabelnas}) have been proven capable of achieving comparable performance to supervised NAS methods. Accordingly, we propose to use an unsupervised learning method as an alternative to supervised distillation in the aforementioned block-wise NAS scheme (Fig. \ref{fig:intro_c}), aiming to address the problem of architectural bias caused by the use of the teacher model. 

In this work, we propose a novel unsupervised NAS method, \textbf{B}l\textbf{o}ck-wisely \textbf{S}elf-\textbf{s}upervised \textbf{N}eural \textbf{A}rchitecture \textbf{S}earch (\textbf{BossNAS}), which aims to address the problem of inaccurate predictive architecture ranking caused by a large weight-sharing space while avoiding possible architectural bias caused by the use of the teacher model. As opposed to the block-wise solutions discussed above, which utilize distillation as intermediate supervision, we propose a self-supervised representation learning scheme named \textbf{ensemble bootstrapping} to optimize each block of our supernet. To be more specific, each sampled sub-networks are trained to predict the \textit{probability ensemble} of all the sampled ones in the target network, between different augmented views of the same image. In the searching stage, an \textbf{unsupervised evaluation metric}, is proposed to ensure fairness by searching towards the architecture population center. More specifically, the probability ensemble of all the architectures in the population is used as the evaluation target to measure the performance of the sampled models.

Additionally, we design a fabric-like \textbf{hy}brid CNN-\textbf{tra}nsformer search space (\textbf{HyTra}) with searchable down-sampling positions and use it as a case study for hybrid architectures to evaluate our method. In each layer of HyTra search space, CNN building blocks and transformer building blocks of different resolutions are in parallel and can be chosen flexibly. This diversified search space covers pure transformers with fixed content length and normal CNNs with progressively reduced spatial scales.

We prove that our NAS method can generalize well on three different search spaces and three datasets. On HyTra search space, our searched models outperforms the ones searched by our supervised NAS counterpart \cite{Li2020Blockwisely_cvpr}, proving that our method successfully avoids possible architecture bias brought by supervised distillation. Our method achieves  superior architecture rating accuracy with 0.78 and 0.76 Spearman correlation on the canonical MBConv search space with ImageNet and on NATS-Bench \textit{size} search space $\mathcal{S_S}$ \cite{dong2021nats} with CIFAR-100, respectively, surpassing \textit{state-of-the-art} NAS methods, proving that our method successfully suppressed the problem of inaccurate architecture rating caused by large weight-sharing space. 

Our searched models on HyTra search space achieves 82.5\% accuracy on ImageNet, surpassing EfficientNet \cite{Tan2019EfficientNetRM} by 2.4\%, with comparable compute time\footnote{Following \cite{srinivas2021bottleneck}, \textit{compute time} refers to the time spent for forward and backward passes.\vspace{-10pt}}. By providing strong results through BossNet-T, we hope that
this diversified HyTra search space with disparate candidates and high-performance architectures can serve as a new arena for future NAS works. We also hope that our BossNAS can serve as a widely used tool for hybrid architecture design.

\vspace{-7pt}
\section{Related Works}
\vspace{-7pt}
\noindent\textbf{Block-wise weight-sharing NAS} \cite{Li2020Blockwisely_cvpr,moons2020donna,zhang2020stagewisepruning,zhang2021semidnagan} approaches factorize the supernet into independently optimized blocks and thus reduce the weight-sharing space, resolving the issue of inaccurate architecture ratings caused by weight-sharing. DNA \cite{Li2020Blockwisely_cvpr} first introduced the block-wisely supervised architecture rating scheme with knowledge distillation. Based on this scheme, DONNA \cite{moons2020donna} further propose to predict an architecture rating using a linear combination of its blockwise ratings rather than a simplistic sum. SP \cite{zhang2020stagewisepruning} were the first to apply this scheme to network pruning.
However, all of the aforementioned methods rely on a supervised distillation scheme, which inevitably introduces architectural bias from the teacher.
We accordingly propose a block-wisely self-supervised scheme, which completely casts off the yoke of the teacher architecture.

\noindent\textbf{Unsupervised NAS} \cite{liu2020unnas,zhang2021randomlabelnas} methods perform architecture search without access to any human-annotated labels. UnNAS \cite{liu2020unnas} introduced unsupervised \textit{pretext tasks} \cite{komodakis2018unsupervised,noroozi2016unsupervised,zhang2016colorful} to weight-sharing NAS for supernet training and architecture rating. RLNAS \cite{zhang2021randomlabelnas} optimized the supernet using \textit{random labels} \cite{zhang2016understanding,maennel2020neural} and further rated architectures by means of a \textit{convergence-based} \textit{angle} metric \cite{hu2020angle}. Another line of NAS methods \cite{yan2020does,wei2020self,hesslow2021contrastive,Peng2021PiNAS} belonging to the category of \textit{supervised NAS} perform unsupervised pretraining of network accuracy predictor or supernet before \textit{supervised} finetuning or evaluation.
Differing from aforementioned works in motivation and methodology, we explore \textit{self-supervised contrastive learning} methods in our unsupervised NAS scheme to avoid the supervision bias in block-wise NAS.

\noindent\textbf{Self-supervised contrastive learning} methods \cite{oord2018representation,wu2018unsupervised,hjelm2018learning,tian2019contrastive,zhuang2019local,he2020momentum,chen2020simple} have significantly advanced the unsupervised learning of visual representations. These approaches learn visual representations in a discriminative fashion by gathering the representations of different views from the same image and spreading those from different images. Recently, the innovative BYOL \cite{grill2020bootstrap} and SimSiam \cite{chen2020exploring} learned visual representations without the use of negative examples. These works directly predict the representation of one view from another using a pair of \textit{Siamese networks} with the same architectures and shared weights \cite{chen2020exploring}, or with one of the Siamese network branches being a momentum encoder, thereby forming a bootstrapping scheme \cite{grill2020bootstrap}. Our work introduces a novel bootstrapping scheme with probability ensemble to \textit{Siamese supernets}.

\noindent\textbf{Architecture Search Spaces.} Cell-based search spaces, first proposed in \cite{zoph2018learning}, are generally used in previous NAS methods \cite{liu2018progressive,Real2019AmoebaNet,Liu2018DARTSDA,pham2018enas} and benchmarks \cite{ying2019bench101,dong2020nasbench201,dong2021nats}. They search for a repeatable cell-level architecture, while keeping a manually designed network-level architecture. By contrast, network-level search spaces with layer-level granularity \cite{Cai2018ProxylessNASDN,Wu2018FBNetHE,chu2021fairnas,Li2020Blockwisely_cvpr,moons2020donna,zhang2021randomlabelnas} and block-level granularity \cite{Tan2018MnasNetPN,Howard2019SearchingFM,Tan2019EfficientNetRM} search for the macro network-level structure using manually designed building blocks (\textit{e.g.} \textit{MBConv} \cite{Sandler2018MobileNetV2IR}). Auto-DeepLab \cite{liu2019auto} presents a hierarchical search space for semantic segmentation, with repeatable cells and a fabric-like \cite{saxena2016convolutional} network-level structure. Our HyTra search space also has a fabric-like network-level structure, albeit with layer-level granularity rather than repeated cells.

\begin{figure*}[t]
\vspace{-20pt}
    \centering
    \includegraphics[width=0.9\linewidth]{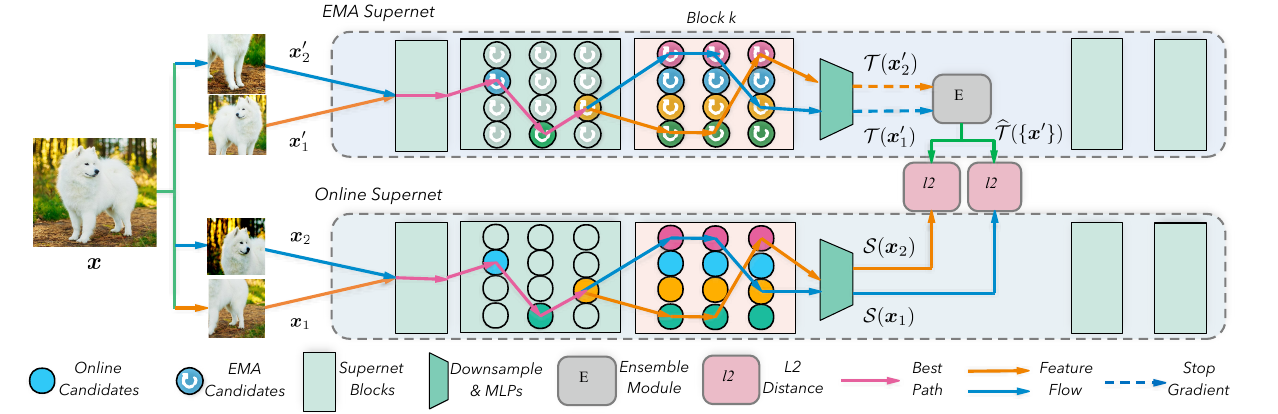}
    \vspace{-5pt}
    \caption{Illustration of the Siamese supernets training with ensemble bootstrapping.
    \vspace{-5pt}}
    \label{fig:Pipline}
\vspace{-10pt}
\end{figure*}
\vspace{-8pt}
\section{Block-wisely Self-supervised NAS}
\vspace{-6pt}
In this section, we first briefly introduce the dilemma of NAS and its block-wise solutions \cite{Li2020Blockwisely_cvpr,moons2020donna,zhang2020stagewisepruning,zhang2021semidnagan}, then present our proposed BossNAS in detail, along with its two key elements: \textbf{i)} unsupervised supernet training phase with \textit{ensemble bootstrapping}; \textbf{ii)} unsupervised architecture rating and searching phase towards architecture population center. 

\noindent\textbf{Notations.} We denote scalars, tensors and sets of tensors using lower case, bold lower case and upper case calligraphic letters respectively (\textit{e.g.}, $n$, $\boldsymbol x$ and $\mathcal{X}$). For simplicity, we use $\{\boldsymbol x_n\}$ to denote the set $\{\boldsymbol x_n\}^{|n|}_{n=1}$ with cardinality $|n|$.
\vspace{-6pt}
\subsection{Dilemma of NAS and the Block-wise Solutions}
\vspace{-6pt}

\noindent\textbf{Dilemma of NAS: efficiency or accuracy.}
While classical sample-based NAS methods produce accurate architecture ratings, they are also computationally prohibitive. Weight-sharing rating scheme in one-shot NAS methods has brought about a tremendous reduction of search cost by encoding the entire search space $\mathcal{A}$ into a weight-sharing supernet, with the weights $\mathcal{W}$ shared by all the candidate architectures
and optimized concurrently as:
\begin{small}$\mathcal{W}^* = \mathop{\arg\min}\limits_\mathcal{W} \boldsymbol{\mathcal{L}}_{\text{train}}(\mathcal{W}, \mathcal{A}; \boldsymbol x, \boldsymbol y)$\end{small}.
Here $\boldsymbol{\mathcal{L}}_{\text{train}}(\cdot)$ denotes the training loss function, while $\boldsymbol x$ and $\boldsymbol y$ denote the input data and the labels, respectively. Subsequently, architectures $\boldsymbol\alpha$ are searched based on the ranking of their ratings with these shared network weights. Without loss of generality, we choose the evaluation loss function $\boldsymbol{\mathcal{L}}_{\text{val}}$ as the rating metric; the searching phase can be formulated as:
\begin{small}$\boldsymbol\alpha^{*} = \mathop{\arg\min}\limits_{\forall \boldsymbol\alpha \in \mathcal{A}}\boldsymbol{\mathcal{L}}_{\text{val}}(\mathcal{W}^{*}, \boldsymbol\alpha; \boldsymbol x, \boldsymbol y)$\end{small}.
However, the architecture ranking based on the shared weights $\mathcal{W}^{*}$ does not necessarily represents the correct ranking of the architectures, as the weights inherited from the supernet are highly entangled and are not fully and fairly optimized. As pointed out in the literature \cite{sciuto2019evaluating,anonymous2020nas,zela2019bench1s1}, weight-sharing methods suffer from low architecture rating accuracy. 

\noindent\textbf{Block-wisely supervised NAS.} As proven theoretically and experimentally by \cite{li2020improving,Li2020Blockwisely_cvpr,moons2020donna}, reducing the weight-sharing space (\textit{i.e.} total number of weight-sharing architectures) can effectively improve the accuracy of architecture rating. In practice, block-wise solutions \cite{Li2020Blockwisely_cvpr,moons2020donna,zhang2020stagewisepruning,zhang2021semidnagan} find a way out of this \textit{dilemma of NAS} by block-wisely factorizing the search space in the depth dimension, thus reducing the \textit{weight-sharing space} while maintaining the original size of the \textit{search space}.
Given a supernet consisting of $|k|$ blocks \begin{small}$\mathcal{S}(\mathcal{W}, \mathcal{A}) = \{\mathcal{S}_k(\mathcal{W}_k, \mathcal{A}_k)\}$\end{small}, with \begin{small}$\mathcal{W} = \{\mathcal{W}_k\}$\end{small} and \begin{small}$\mathcal{A} = \{\mathcal{A}_k\}$\end{small} denoting its weights and architecture that are block-wisely separable in the depth dimension, each block of the supernet is trained separately before searching among all blocks in combination by the sum \cite{Li2020Blockwisely_cvpr}, or a linear combination \cite{moons2020donna} (with weights $\{\lambda_k\}$), of each block's evaluation loss $\boldsymbol{\mathcal{L}}_{\text{val}}$:
\vspace{-8pt}
\begin{equation}\label{eq:loss-dna}\small
\begin{aligned}\
&\boldsymbol\alpha^{*} = \{\boldsymbol\alpha_k\}^{*} = \mathop{\arg\min}\limits_{\forall \{\boldsymbol\alpha_k\} \subset \mathcal{A}}\sum_{k=1}^{|k|} \lambda_k \boldsymbol{\mathcal{L}}_{\text{val}}\Big(\mathcal{W}_k^{*},\boldsymbol\alpha_k; \boldsymbol x_k, \boldsymbol y_k\Big) \\\vspace{-5pt}
&s.t. ~~\mathcal{W}_k^{*} = \mathop{\arg\min}_{\mathcal{W}_k} \boldsymbol{\mathcal{L}}_{\text{train}}(\mathcal{W}_k, \mathcal{A}_k; \boldsymbol x_k, \boldsymbol y_k)
\end{aligned}.
\vspace{-7pt}
\end{equation}

To isolate the training of each supernet block, given an input $\boldsymbol x$, the intermediate input and target $\{\boldsymbol x_k,\boldsymbol y_k\}$ of the $k$-th block is generated by a fixed teacher network $\mathcal{T}$ (with architecture $\mathcal{\boldsymbol\alpha}^\mathcal{T}$ and ground-truth weights $\mathcal{W}^\mathcal{T}$): \begin{small}$\{\boldsymbol x_1,\boldsymbol y_1\} = \{\boldsymbol x, \mathcal{T}_1(\boldsymbol x)\}$\end{small}, and \begin{small}$\{\boldsymbol x_k,\boldsymbol y_k\} = \{\mathcal{T}_{k-1}(\boldsymbol x), \mathcal{T}_k(\boldsymbol x)\}, k>1$\end{small}, where $\mathcal{T}_k$ represents the teacher network truncated after the $k$-th block.
As the data used for both training and searching phase are generated by the teacher model \begin{small}$\mathcal{T}(\mathcal{W}^\mathcal{T}, \mathcal{\boldsymbol\alpha}^\mathcal{T})$\end{small}, the architecture ratings are likely to be highly correlated with the teacher architecture. For instance, a convolutional teacher have a \textit{limited receptive field} and distinctive architectural inductive biases like \textit{translation equivariance}. With such a biased supervision, candidate architectures are likely to be trained and rated unfairly. We observes two phenomenons that can be attrbute to the biased supervision, \textit{i.e.} \textit{candidate preference} and \textit{teacher preference}.
Detailed experimental analysis of these two phenomenons is provided in Sec. \ref{sec:exp}. To break these restrictions of current block-wise NAS solutions, we explore a scheme without using a teacher model.
\begin{figure*}[ht]
\vspace{-15pt}
    \centering
    \includegraphics[width=0.9\linewidth]{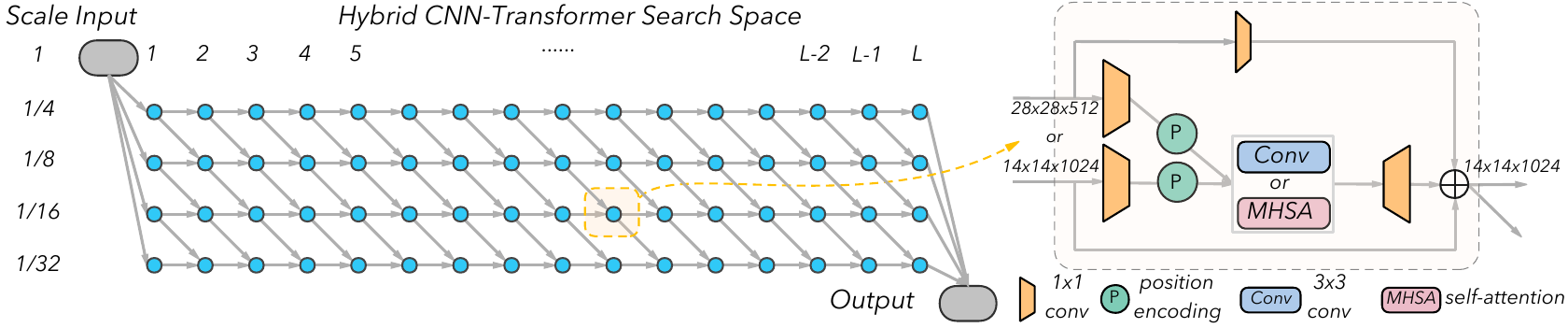}
    \vspace{-5pt}
    \caption{Illustration of the fabric-like Hybrid CNN-transformer Search Space with flexible down-sampling positions.}
    \vspace{-12pt}
    \label{fig:SearchSpaceT}
\end{figure*}
\vspace{-7pt}
\subsection{Training with Ensemble Bootstrapping}
\vspace{-5pt}

Starting from the dual network scheme with student-teacher pair $\{\mathcal{S}(\mathcal{W}, \mathcal{A}), \mathcal{T}(\mathcal{W}^\mathcal{T}, \boldsymbol\alpha^\mathcal{T})\}$, the first step to cast off the yoke of the teacher architecture is to assign $\boldsymbol\alpha^\mathcal{T} = \mathcal{A}$, thus forming a pair of \textit{Siamese supernets}.

\noindent\textbf{Bootstrapping with Siamese Supernets.}
To optimize such Siamese networks block-wisely, we adopt a self-supervised contrastive learning scheme. More specifically, these two supernets receive a pair of augmented views $\{\boldsymbol x_1, \boldsymbol x_2\}$ of the same training sample $\boldsymbol x$ and generate the outputs 
$\small\{\mathcal{S}(\mathcal{W}, \mathcal{A};\boldsymbol x_1), \mathcal{T}(\mathcal{W}^\mathcal{T}, \mathcal{A};\boldsymbol x_2)\}$, respectively. Analogous to previous teacher-student settings, the Siamese supernets are optimized by minimizing the distance between their outputs. In previous Siamese networks and self-supervised contrastive learning methods, the two networks either share their weights \cite{chen2020simple,chen2020exploring} (\textit{i.e.} $\mathcal{W}^\mathcal{T} = \mathcal{W}$) or form a mean teacher scheme with Exponential Moving Average (EMA) \cite{he2020momentum,grill2020bootstrap} (\textit{i.e.} $\mathcal{W}^\mathcal{T} = \mathcal{W}^\bullet$, where $\mathcal{W}^\bullet_t = \tau \mathcal{W}^\bullet_{t-1} + (1-\tau)\mathcal{W}_t$ represents the temporal average of $\mathcal{W}$, with $t$ being a training timestamp, and $\normalsize\tau$ denoting the momentum factor that controls the updating speed of $\mathcal{W}^\bullet$). By learning representation from the mean teacher, analogous to the simple yet powerful BYOL \cite{grill2020bootstrap}, our supernet can be optimized in an unsupervised manner without relying on a fully supervised teacher network:\vspace{-8pt}
\begin{small}\begin{equation}\label{eq:loss-boss}
\mathcal{W}_k^{*} = \mathop{\arg\min}_{\mathcal{W}_k} \boldsymbol{\mathcal{L}}_{\text{train}}\Big(\{\mathcal{W}_k,\mathcal{W}_k^\bullet\}, \mathcal{A}_k; \boldsymbol x_k\Big).
\vspace{-7pt}\end{equation}\end{small}
To eliminate the influence of pixel-wise differences between two intermediate representations caused by augmentations (\textit{e.g.} random crop), as well as to ensure better generalization on candidate architectures with different reception fields or even different resolutions, we project the representations to the latent space before calculating the element-wise distance.

\noindent\textbf{Ensemble Bootstrapping.}
However, unlike single networks, supernets are typically optimized by path sampling strategies, \textit{e.g.} single path \cite{Guo2019SinglePO} or fair path \cite{chu2021fairnas}. When naively adopting bootstrapping, each sub-network learns from the moving average of itself. In the absence of a common objective, the weights shared by different sub-networks suffer from convergence hardship, leading to training instability and inaccurate architecture ratings. To address this problem, we propose an unsupervised supernet training scheme, named \textit{ensemble bootstrapping}.

Considering $|p|$ sub-networks \begin{small}$\{\boldsymbol\alpha_p\} \subset \mathcal{A}_k$\end{small} sampled from the $k$-th block of the search space $\mathcal{A}$ in the $t$-th training iteration, and given a training sample $\boldsymbol x$, $|p|$ pairs of augmented views \begin{small}$\{\boldsymbol x_p\} \sim \boldsymbol p_{aug}(\cdot|\boldsymbol x)$\end{small}, \begin{small}$\{\boldsymbol x'_p\} \sim \boldsymbol p'_{aug}(\cdot|\boldsymbol x)$\end{small} are generated for each sampled sub-network of the Siamese supernets. To form a common objective for all paths, we can use a scheme analogous to ensemble distillation \cite{Meal, Mealv2} in supervised learning. As illustrated in Fig. \ref{fig:Pipline}, each sampled sub-network of the online supernet learns to predict the \textit{probability ensemble} of all sampled sub-networks in the EMA supernet:
\vspace{-10pt}\begin{equation}\label{eq:ensemble-probability}
\small
\begin{aligned}
    \widehat{\mathcal{T}_k}\Big(
\{\boldsymbol\alpha_p\}; \{\boldsymbol x'_p\}\Big) = \frac{1}{|p|} \sum_{p=1}^{|p|}\mathcal{T}_{k}(\mathcal{W}^\bullet, \boldsymbol\alpha_p;\boldsymbol x'_p).
\end{aligned}
\vspace{-2pt}\end{equation}
In summary, the block-wisely self-supervised training process of the Siamese supernets is formulated as follows:
\vspace{-6pt}\begin{equation}\label{eq:train_boss}
\small
\begin{aligned}
\mathcal{W}_k^{*} &= \mathop{\arg\min}_{\mathcal{W}_k} \sum\limits^{|p|}_{p=1}\boldsymbol{\mathcal{L}}_{\text{train}}\Big(\{\mathcal{W}_k,\mathcal{W}_k^\bullet\}, \{\boldsymbol\alpha_p\}; \boldsymbol x\Big), \\[-2pt]
\text{where~~~~} &\boldsymbol{\mathcal{L}}_{\text{train}}\Big(\{\mathcal{W}_k,\mathcal{W}_k^\bullet\}, \{\boldsymbol\alpha_p\}; \boldsymbol x\Big)\\[-2pt]
= &\left\|\mathcal{S}_k(\mathcal{W}_k,
\boldsymbol\alpha_p; \boldsymbol x_p) -  \widehat{\mathcal{T}_k}\Big(\mathcal{W}^{\bullet}_k,
\{\boldsymbol\alpha_p\}; \{\boldsymbol x'_p\}\Big)\right\|_2^2.
\end{aligned}\vspace{-8pt}
\end{equation}

\vspace{-4pt}
\subsection{Searching Towards the Population Center}
\vspace{-5pt}
After the convergence of the Siamese supernets is complete, the architectures can be ranked and searched by the rating determined based on the weight of the supernets, as in Eqn. \ref{eq:loss-dna}. In this section, we design a fair and effective unsupervised rating metric $\boldsymbol{\mathcal{L}}_{\text{val}}$ for searching phase.

To evaluate the performance of a network trained with contrastive self-supervision, previous works \cite{he2020momentum,chen2020simple,grill2020bootstrap,chen2020exploring} have utilized supervised metrics, such as accuracies of linear evaluation or few-shot classification.
To develop an unsupervised NAS method, we aim to avoid schemes that depend on human-annotated labels and instead pursue a completely unsupervised evaluation metric. Previous unsupervised NAS methods \cite{liu2020unnas,zhang2021randomlabelnas} utilize either the accuracy of pretext tasks or convergence measurement with angle-based metrics to rate candidate architectures. Unfortunately, the losses of self-supervised contrastive learning do not necessarily represent either the architecture performance or the architecture convergence, as the input views and target networks are both \textit{randomly sampled}. Moreover, the target networks are somewhat biased and cannot serve as ground truth targets. To avoid these concerns, we propose a fair and effective unsupervised evaluation metric for architecture search. 

Without loss of generality, we consider searching with an \textit{evolutionary algorithm} \cite{chen2018reinforced,Real2019AmoebaNet}, where architectures are optimized by evolving an architecture population $\{\boldsymbol\alpha_p\}$.
Analogous to the optimization of the weights, we propose to use \textit{probability ensemble} among the population $\{\boldsymbol\alpha_p\}$ as the common target to provide a fair rating for each architecture $\boldsymbol\alpha_p$. Additionally, one pair of views $\{\boldsymbol x_1, \boldsymbol x_2\}$ for each validation samples $\boldsymbol x$ are generated and \textit{fixed} to avoid the bias introduced by variable augmentation. In parallel to Eqn. \ref{eq:ensemble-probability}, we have the probability ensemble of the architecture population:
\vspace{-10pt}\begin{equation}\label{eq:ensemble-probability-search}
\small
\begin{aligned}
    \widehat{\mathcal{S}_k}\Big(
\{\boldsymbol\alpha_p\}; \boldsymbol x_2\Big) = \frac{1}{|p|} \sum_{p=1}^{|p|}\mathcal{S}_{k}(\boldsymbol\alpha_p;\boldsymbol x_2).
\end{aligned}
\vspace{-4pt}\end{equation}

In practice, by dividing the supernet into medium-sized blocks (\textit{e.g.} 4 layers of 4 candidates, $4^4=256$ architectures), traversal evaluation of all the candidate architectures are affordable.
In this case, the architecture population $\{\boldsymbol\alpha_p\}$ is expanded to the whole block-wise search space $\mathcal{A}_k$, and the whole searching process is finished in a single step:
\vspace{-7pt}\begin{equation}\label{eq:search_boss}
\small
\begin{aligned}
&\boldsymbol\alpha^{*} = \mathop{\arg\min}\limits_{\forall \boldsymbol\alpha \in \mathcal{A}}\sum_{k=1}^{|k|} \lambda_k \boldsymbol{\mathcal{L}}_{\text{val}}(\boldsymbol\alpha; \boldsymbol x_k)\\[-2pt]
\text{where~~~~} &\boldsymbol{\mathcal{L}}_{\text{val}}(\boldsymbol\alpha; \boldsymbol x) = \left\|\mathcal{S}_k(
\boldsymbol\alpha; \boldsymbol x_1) -  \widehat{\mathcal{S}}_k(
\mathcal{A}_k; \boldsymbol x_2)\right\|_2^2.
\end{aligned}
\vspace{-6pt}\end{equation}

\vspace{-6pt}
\section{Hybrid CNN-transformer Search Space}\label{sec:searchspaceT}
\vspace{-5pt}
In this section, we present a fabric-like hybrid CNN-transformer search space, named HyTra, with disparate candidate building blocks and flexible down-sampling positions.%
\vspace{-5pt}
\subsection{CNN and Transformer Candidate Blocks}
\vspace{-5pt}
The first step in designing a hybrid CNN-transformer search space is to include the proper CNN and transformer building blocks. These two types of building blocks should be able to perform well either when simply aggregated in sequence or when combined freely. We choose the classical and robust \textit{residual bottleneck} ({\tt \textbf{ResConv}}) in ResNet \cite{he2016resnet} as the CNN candidate building block. In parallel, we design a lightweight and robust transformer building block {\tt \textbf{ResAtt}} based on the pluggable \textit{BoTBlock} \cite{srinivas2021bottleneck} and \textit{NLBlock} \cite{wang2018non}.

\begin{figure}[t]
    \centering
    \includegraphics[width=0.9\linewidth]{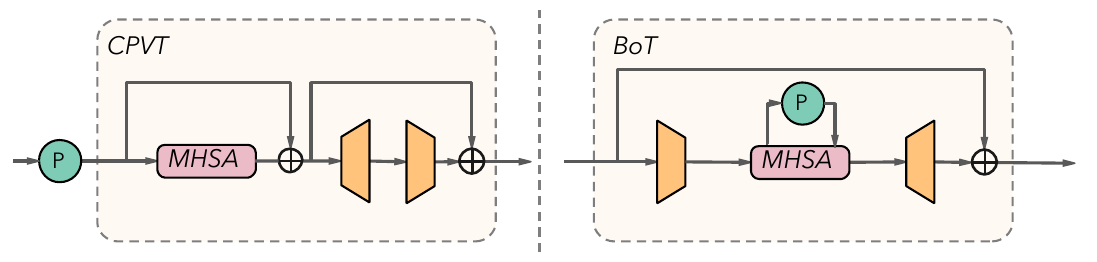}
    \vspace{-10pt}
    \caption{Transformer blocks in CPVT \cite{chu2021conditional} and BoT \cite{srinivas2021bottleneck}.}
    \vspace{-10pt}
    \label{fig:CPVTBlock}
    \vspace{-5pt}
\end{figure}

\noindent\textbf{Computation Balancing with Implicit Position Encodings.}
To facilitate fair and meaningful competition, candidate building blocks should have similar computation complexities.
The original BoTBlock is slower than {\tt \textbf{ResConv}}, as its relative position encodings are computed separately through multiplication with the \textit{query}. Simply removing the content-position branch from BoTBlocks, resembling to NLBlocks, could reduce their compute time to make them comparable to {\tt \textbf{ResConv}}. However, position encodings are crucial for vision transformers to achieve good performance.
In CPVT \cite{chu2021conditional}, the authors uses single convolutions in between transformer encoder blocks as the \textit{position encoding generator}. Similarly, we replace the relative position encoding branch in BoTBlock with a light depthwise separable convolution as an implicit position encoding module, forming our {\tt \textbf{ResAtt}}. By this simple modification, we reduce the computation complexity of position encoding module from the original $\mathcal{O}(CW^3)$ to $\mathcal{O}(CW^2)$, with $C$ denoting number of channels and $W$ denoting the width or height. In contrast to CPVT and BoT (Fig. \ref{fig:CPVTBlock}), our position encoding modules (Fig. \ref{fig:SearchSpaceT} right) are placed between the input projection layer and the self-attention module. In addition, our implicit position encoding modules are also responsible for down-sampling. This modification is also applied to {\tt \textbf{ResConv}}, which enables weight sharing between candidate blocks with different down-sampling rates (\textit{i.e.} 1 or 2). 

\vspace{-5pt}
\subsection{Fabric of Hybrid CNN-transformers}
\vspace{-5pt}
Beyond the building blocks, CNNs and transformers differ considerably in terms of their macro architectures. Unlike CNNs, which process images in stages with various spatial sizes, transformers typically do not change sequence length (image patches) and retains the same scale at each layer. As shown in Fig. \ref{fig:SearchSpaceT} left, to cover both the CNNs and transformers, our search space is designed with flexible down-sampling positions, forming a \textit{fabric} \cite{saxena2016convolutional} of Hybrid CNN-transformers. At each choice block layer of the fabric, the spatial resolution can either stay unchanged or be reduced to half of its scale, until reaching the smallest scale.
This fabric-like search space contains architectures resembling the popular vision transformers \cite{dosovitskiy2021vit,touvron2020deit,chu2021conditional}, CNNs \cite{he2016resnet,Hu2019SqueezeandExcitationN} and hybrid CNN-transformers \cite{srinivas2021bottleneck} at different scales.
\begin{table}
\vspace{-6pt}
    \scriptsize
    \centering
    \begin{tabular}{l|c|c|c|c}
    \toprule
    Method      & MAdds & Steptime & Top-1 (\%) &Top-5 (\%)\\
    \midrule
    ResNet50 \cite{he2016resnet}         & 4.1B & 100ms & 77.7 & 93.9\\
    ViT-B/32 \cite{dosovitskiy2021vit}  & - & 68ms & 73.4 & - \\
    ViT-B/16 \cite{dosovitskiy2021vit}  & 17.6B & 158ms  & 77.9& -\\
    BoT50 \cite{srinivas2021bottleneck}     & 4.0B & 120ms & 78.3& 94.2\\
    \arrayrulecolor{lightgray}\hline\arrayrulecolor{black}
    \textcolor{RoyalPurple}{R50-T ~~~~~~$\mathtt{Conv}$-$\mathtt{Only}$} & 4.1B & 104ms & 78.2&94.2\\
    \textcolor{RoyalPurple}{ViT-T/32 ~~~$\mathtt{Att}$-$\mathtt{Only}$} & 2.9B & 92ms & 74.5 & 91.7 \\
    \textcolor{RoyalPurple}{ViT-T/16 ~~~$\mathtt{Att}$-$\mathtt{Only}$} & 3.2B & 96ms & 76.5& 93.0\\
    \textcolor{RoyalPurple}{BoT50-T ~~~$\mathtt{Hybrid}$} & 3.9B & 103ms & 79.5 & 94.8 \\
    \textcolor{RoyalPurple}{Random-T ~$\mathtt{Hybrid}$} & 3.7B  & 84ms  & 76.7 & 93.1 \\
    \arrayrulecolor{lightgray}\hline\arrayrulecolor{black}
    \textbf{BossNet-T0 $\mathtt{~w/o~SE}$} & 3.4B & 101ms &\textbf{80.5}& \textbf{95.0}\\
    \midrule
    SENet50  \cite{he2016resnet,Hu2019SqueezeandExcitationN}    & 4.1B & 129ms &79.4&94.6\\
    EffNetB1 \cite{Tan2019EfficientNetRM}    & 0.7B & 131ms & 79.1 & 94.4\\
    DeiT-S  \cite{touvron2020deit} & 10.1B & 84ms & 79.8 & - \\
    BoT50 + SE \cite{srinivas2021bottleneck}  & 4.0B & 149ms & 79.6 & 94.6\\
    \arrayrulecolor{lightgray}\hline\arrayrulecolor{black}
    DNA-T  \cite{Li2020Blockwisely_cvpr} & 3.9B & 121ms & 80.3 & 95.0\\
    UnNAS-T \cite{liu2020unnas} &  3.7B & 104ms &79.8&94.6\\
    \arrayrulecolor{lightgray}\hline\arrayrulecolor{black}
    \textbf{BossNet-T0} & 3.4B & 115ms & 80.8 & 95.2\\
    \textbf{BossNet-T0} $\boldsymbol\uparrow$ & 5.7B & 147ms & \textbf{81.6} & \textbf{95.6}\\
    \midrule
    SENet101  \cite{he2016resnet,Hu2019SqueezeandExcitationN}& 7.8B & 218ms &81.4&95.7\\
    EffNetB2 \cite{Tan2019EfficientNetRM} & 1.0B & 143ms & 80.1 & 94.9 \\
    ViT-L/16 \cite{dosovitskiy2021vit} & 63.6B &  168ms & 81.1& -\\
    DeiT-B \cite{touvron2020deit} & 17.6B & 152ms & 81.8 & -\\
    BoTNet-S1-59 \cite{srinivas2021bottleneck}  &7.3B & 184ms & 81.7 &95.8\\
    T2T-ViT-19 \cite{yuan2021t2t}& 8.9B & 158ms & 81.9& -\\
    TNT-S \cite{han2021tnt}& 5.2B & 468ms &81.3 &95.6\\
    \arrayrulecolor{lightgray}\hline\arrayrulecolor{black}
    \textbf{BossNet-T1} & 7.9B & 156ms & 82.2 & 95.8\\
    \textbf{BossNet-T1} $\boldsymbol\uparrow$ & 10.5B & 165ms & \textbf{82.5} & \textbf{96.0} \\
    \bottomrule
    \end{tabular}
    \vspace{-5pt}
    \caption{\textbf{ImageNet} results of \textit{state-of-the-art} models and our searched \textbf{hybrid CNN-transformers}. Compute steptime is measured on a single GeForce RTX 3090 GPU with batch size 32. \textcolor{RoyalPurple}{Purple} is used to denote manually selected architectures from search space HyTra. $\normalsize\boldsymbol\uparrow$: Directly tested on larger input size without finetuning (\textit{i.e.} 288 for BossNet-T0$\normalsize\boldsymbol\uparrow$ and 256 for BossNet-T1$\normalsize\boldsymbol\uparrow$).}
    \label{tab:T_acc_r50}
    \vspace{-15pt}
\end{table}
\vspace{-5pt}
\section{Experiments}\label{sec:exp}
\vspace{-5pt}
\noindent\textbf{Setups.}
We evaluate our method on three search spaces, including our proposed HyTra search space and other two existing search spaces,\textit{ i.e.} MBConv search space \cite{Cai2018ProxylessNASDN,Li2020Blockwisely_cvpr} and NATS-Bench \textit{size} search space $\mathcal{S_S}$ \cite{dong2021nats}. The datasets we use to evaluate and analyze our method are ImageNet \cite{deng2009imagenet},
CIFAR-10 and CIFAR-100 \cite{Krizhevsky09cifar}.
We train each block of the supernet for 20 epochs, including one linear warm-up epoch. We randomly sample four paths in each training step. 
See Appendix \ref{sec:app_details} for more implementation details.

\vspace{-5pt}
\subsection{Searching for Hybrid CNN-transformer}
\vspace{-5pt}
\noindent\textbf{Analysis of HyTra search space.}\label{sec:HyTra_ablation} We manually stitched four architectures on our fabic-like HyTra search space, following as closely as possible to previous human-designed networks \cite{he2016resnet,dosovitskiy2021vit,srinivas2021bottleneck}, except using our $\normalsize\{\mathtt{ResConv, ResAtt}\}$ building blocks.
As shown in Tab. \ref{tab:T_acc_r50}, these models (in \textcolor{RoyalPurple}{purple}) consistently outperform their prototypes. Remarkably, BoT50-T surpasses the original BoT50 by \textbf{1.2\%} top-1 accuracy with \textbf{1.17$\times$} compute time reduction, proving the superiority of our designed building blocks.

\begin{figure}
\vspace{-8pt}
\centering
\begin{subfigure}[t]{\linewidth}
    \centering\includegraphics[width=0.92\linewidth]{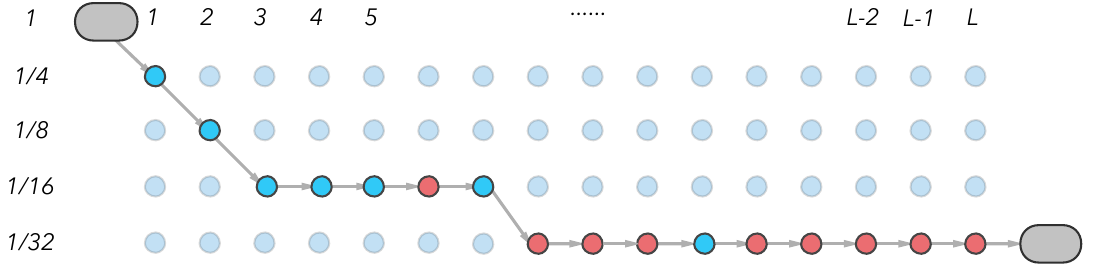}\vspace{-4pt}
    \caption{\small Architecture of BossNet-T0.}\label{fig:BossT}
  \end{subfigure}\\\vspace{0pt}
  \begin{subfigure}[t]{\linewidth}
    \centering\includegraphics[width=0.92\linewidth]{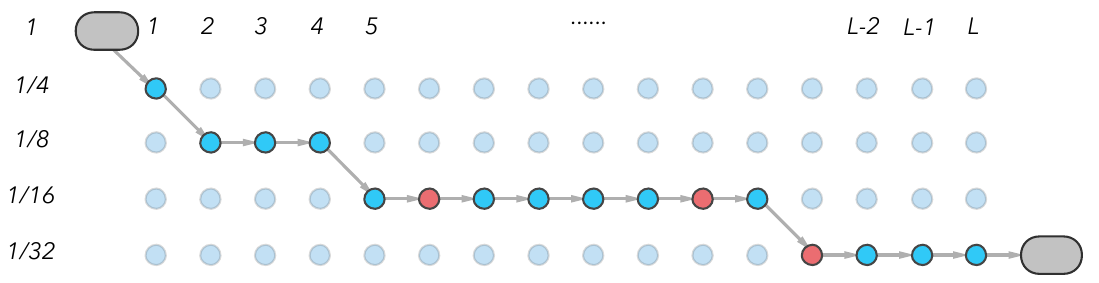}\vspace{-4pt}
    \caption{\small Architecture of DNA-T.}\label{fig:DNAT}
  \end{subfigure}\\
  \vspace{-5pt}
  \caption{Visualization of architectures searched by BossNAS and DNA \cite{Li2020Blockwisely_cvpr} in HyTra search space. \textcolor{cyan}{Blue} nodes denotes \textcolor{cyan}{{\tt \textbf{ResConv}}} and \textcolor{red}{red} nodes denotes \textcolor{red}{{\tt \textbf{ResAtt}}}.}\label{fig:architecture_visualize}
\vspace{-20pt}
\end{figure}
\noindent\textbf{Performance of searched models.} With our proposed search space and NAS method, we explore hybrid CNN-transformer architectures on ImageNet. The results of our searched models (BossNet-T) and models with comparable compute time are summarized in Tab. \ref{tab:T_acc_r50}. 

\textbf{Firstly}, BossNet-T0 outperforms a wide range of \textit{state-of-the-art} models. For instance, BossNet-T0 without SE module achieves \textbf{80.5\%} top-1 accuracy, surpassing the human-designed hybrid CNN-transformer, BoTNet50, by \textbf{2.2\%} while being \textbf{1.19$\times$} faster in terms of compute time; when equipped with SE and SiLU activation, BossNet-T0 further achieves \textbf{80.8\%} top-1 accuracy, surpassing the NAS searched EfficientNet-B1 by \textbf{1.7\%} while being \textbf{1.14$\times$} faster.

\textbf{Secondly}, our searched model demonstrates absolute superiority over manually and randomly selected models from search space HyTra. In particular, BossNet-T0 achieves up to \textbf{6.0\%} improvement over manually selected models, proving the effectiveness of our architecture search.

\textbf{Thirdly}, BossNet-T0 outperforms other recent NAS methods on search space HyTra. BossNet-T0 achieves \textbf{0.5\%} accuracy gains over DNA-T, which is searched by our supervised NAS counterpart \cite{Li2020Blockwisely_cvpr}.

\textbf{Finally}, when extended to larger model size and input size, the family of BossNet-T models maintain their superiority. By removing the downsampling in the last stage of BossNet-T0 (same scheme as BoTNet-S1 \cite{srinivas2021bottleneck}), we have BossNet-T1, which achieves \textbf{82.2\%} accuracy, surpassing EfficientNet-B2 by \textbf{2.1\%}. By directly testing on larger input resolutions without finetuning, BossNet-T0$\normalsize\boldsymbol\uparrow$ (on 288$\times$288 input size) achieves \textbf{81.6\%} top-1 accuracy, and outperforms BoTNet50 + SE by \textbf{2.0\%} with similar runtime; BossNet-T1$\normalsize\boldsymbol\uparrow$ (on 256$\times$256 input size) achieves \textbf{82.5\%} top-1 accuracy, surpassing T2T-ViT-19 and EfficientNet-B2 by \textbf{0.6\%} and \textbf{2.4\%} with comparable steptime, respectively.

\begin{figure*}
\vspace{-16pt}
\centering
\begin{subfigure}[t]{0.83\linewidth}\flushleft
\includegraphics[width=\linewidth]{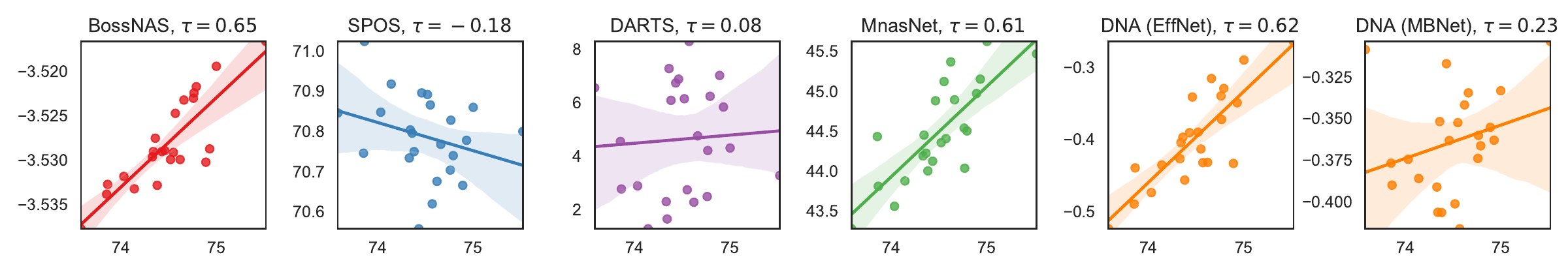}%
\end{subfigure}~~~~%
\begin{subfigure}[t]{0.14\linewidth}
\includegraphics[width=\linewidth]{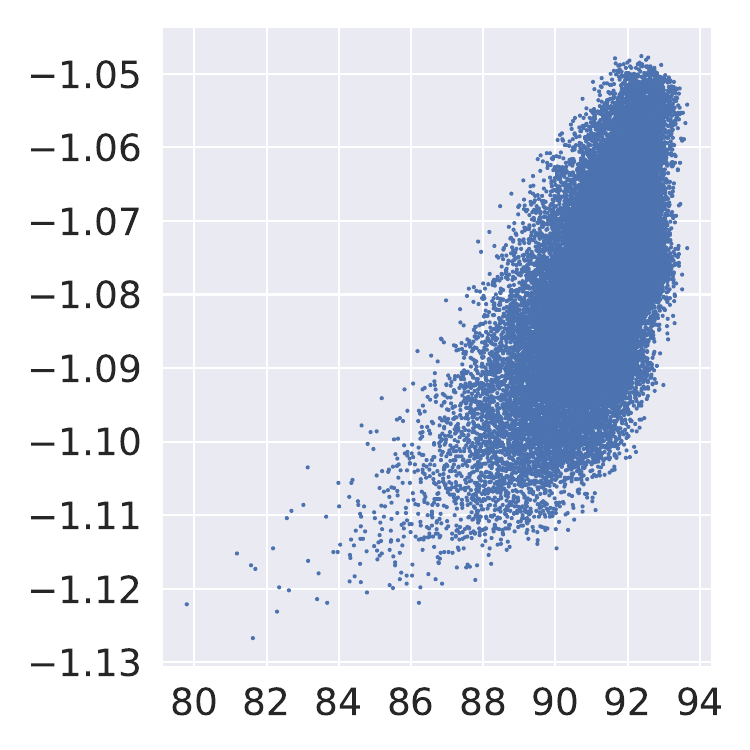}%
\end{subfigure}%
\vspace{-10pt}
\caption{\textbf{Left:} Ranking correlations of 6 different NAS methods on \textbf{MBConv Search Space}.
\textbf{Right:} Architecture ranking of BossNAS on \textbf{NATS-Bench} $\boldsymbol{\mathcal{S_S}}$. In all the diagrams, x-axis denotes ground truth accuracy; y-axis denotes evaluation metrics.
}\label{fig:MB_rank}\vspace{-5pt}
\end{figure*}

\noindent\textbf{Architecture visualization and analysis.}\label{sec:architecture_visualization} We visualize the architecture of DNA-T and BossNet-T0 in Fig. \ref{fig:architecture_visualize}. DNA-T clearly prefers convolutions, as it contains 13 {\tt \textbf{ResConv}} blocks and only three {\tt \textbf{ResAtt}} blocks. By contrast, BossNet-T0 has similar numbers of convolutions and attentions and eventually achieves a higher accuracy. 
We refer this to \textbf{\textit{Phenomenon I: candidate preference}}, and attribute it to architectural bias from the teacher supervision. Without using the teacher model, our method successfully avoids this bias.

\begin{table}
\vspace{-10pt}
    \footnotesize
    \centering
    \begin{tabular}{l|c|c|c}
    \toprule
    Method & MAdds (M) & Top-1 (\%) & Top-5 (\%)\\
    \midrule
    FairNAS-A \cite{chu2021fairnas} & 388M & 75.3 & 92.4\\
    ProxylessNAS \cite{Cai2018ProxylessNASDN} & 465M & 75.1 & 92.5\\
    FBNet-C \cite{Wu2018FBNetHE} & 375M & 74.9 & -\\
    SPOS \cite{Guo2019SinglePO} & 472M & 74.8 & -\\
    RLNAS \cite{zhang2021randomlabelnas} & 473M & 75.6 & 92.6\\
    \textbf{BossNet-M1} $\mathtt{~w/o~SE}$ & 475M & \textbf{76.2} & \textbf{93.0}\\
    \midrule
    MobileNetV3 \cite{Howard2019SearchingFM} & 219M & 75.2 & -\\
    MnasNet-A3 \cite{Tan2018MnasNetPN} & 403M & 76.7 & 93.3\\
    EfficientNet-B0 \cite{Tan2019EfficientNetRM} & 399M & 76.3 & 93.2\\
    DNA-b \cite{Li2020Blockwisely_cvpr} & 406M & \textbf{77.5} & 93.3\\
    \textbf{BossNet-M2} & 403M& 77.4 & \textbf{93.6}\\
    \bottomrule
    \end{tabular}
    \vspace{-8pt}
    \caption{\textbf{ImageNet} results of \textit{state-of-the-art} NAS models on \textbf{MBConv search space}.
    }
    \label{tab:M_acc}
    \vspace{-10pt}
\end{table}

\begin{table}
    \footnotesize
    \centering
    \begin{tabular}{l|c|c|c|c}
        \toprule
        Method & Search Cost & $\normalsize\tau$  & $\normalsize\rho$ & $\normalsize R$ \\\midrule
        SPOS \cite{Guo2019SinglePO} & 8.5 Gds & -0.18 & -0.27 & -0.29  \\
        DARTS \cite{Liu2018DARTSDA} & 50 Gds & 0.08 & 0.14 & 0.06 \\
        MnasNet \cite{Tan2018MnasNetPN} & 288 Tds & 0.61 & \underline{0.77} & 0.78 \\
        DNA \cite{Li2020Blockwisely_cvpr} (EffNetB0) & 8.5 Gds & \underline{0.62} & \underline{0.77} & \underline{0.83} \\
        DNA  \cite{Li2020Blockwisely_cvpr} (MBNetV1) & 8.5 Gds & 0.23 & 0.27 & 0.37 \\
        \textbf{BossNAS} & 10 Gds & \textbf{0.65} & \textbf{0.78} & \textbf{0.85}\\
        \bottomrule
    \end{tabular}
    \vspace{-8pt}
    \caption{Comparison of the effectiveness and efficiency of different NAS methods on \textbf{MBConv search space} and \textbf{ImageNet dataset}.
    (Gds: GPU days; Tds: TPU days)}
    \label{tab:MB_rank}
    \vspace{-15pt}
\end{table}

\vspace{-10pt}
\subsection{Results on MBConv Search Space}
\vspace{-5pt}
To further prove the effectiveness and generalization ability of BossNAS, we compare it with a wide range of NAS methods on MBConv search space. 

\noindent\textbf{Performance of searched models.} As shown in Tab. \ref{tab:M_acc}, our searched models, BossNet-M, achieve competitive results in search spaces with and without SE module. In the search space \textit{without} SE, BossNet-M1, searched under constraint of 475M MAdds, outperforms SPOS \cite{Guo2019SinglePO} and another recent unsupervised NAS method, RLNAS \cite{zhang2021randomlabelnas} by \textbf{1.4\%} and \textbf{0.6\%}, respectively. In the search space \textit{with} SE, BossNet-M2, under constraint of 405M MAdds, outperforms the popular EfficientNet \cite{Tan2019EfficientNetRM} by \textbf{1.1\%}, and is also competitive with our supervised counterpart, DNA \cite{Li2020Blockwisely_cvpr}. Note that candidate building blocks in MBConv search space are quite similar, concealing the \textit{candidate preference} phenomenon in \cite{Li2020Blockwisely_cvpr}.

\noindent\textbf{Architecture rating accuracy.}
As BossNAS performs \textit{traversal search} (\textit{i.e.} accuracy of searching phase is \textbf{100\%}), the \textit{architecture rating accuracy} directly represents its effectiveness. We use the 23 open-sourced architectures in MBConv search space and their corresponding ground truth accuracies provided by \cite{Li2020Blockwisely_cvpr} to calculate the architecture rating accuracy, \textit{i.e.} the ranking correlation between the predicted architecture ranking and the ground truth model ranking. We use three different ranking correlation metrics: Kendall Tau ($\normalsize\tau$) \cite{kendall1938new}, Spearman Rho ($\normalsize\rho$) and Pearson R ($\normalsize R$). All three metrics range from -1 to 1, with ``-1'' representing a completely reversed ranking, ``1'' meaning an entirely correct ranking, and ``0'' representing no correlation between rankings. As shown in Tab. \ref{tab:MB_rank} and Fig. \ref{fig:MB_rank} left, our BossNAS obtains the highest rating accuracies with \textbf{0.65} $\normalsize\boldsymbol\tau$ among \textit{sota} NAS methods, while addressing two problems. 

\textbf{First}, classic weight sharing methods, SPOS \cite{Guo2019SinglePO} and DARTS \cite{Liu2018DARTSDA}, fails to achieve reasonable ranking correlation despite their lower search costs, while the multi-trial method, MnasNet \cite{Tan2018MnasNetPN}, achieves high rating accuracies with massive search cost. BossNAS successfully addressed such \textbf{\textit{dilemma of NAS}} by achieving even higher rating accuracies than MnasNet (\textit{e.g.} \textbf{0.07 $\normalsize \boldsymbol R$}) with \textbf{28.8$\times$} acceleration. 

\textbf{Second}, supervised block-wise NAS method, DNA \cite{Li2020Blockwisely_cvpr}, fails to achieve high rating accuracies when using a teacher largely different from the candidates (MobileNetV1 \cite{howard2017mobilenets} \textit{vs.} EfficientNet-based candidates \cite{Tan2019EfficientNetRM}), which we refer to as \textbf{\textit{Phenomenon II: teacher preference}}. Our unsupervised BossNAS achieves higher rating accuracies than DNA (\textbf{0.03 $\normalsize\boldsymbol\tau$}), successfully casting off the yoke of the teacher network.

\begin{table}
\vspace{-10pt}
    \footnotesize
    \centering
    \setlength{\tabcolsep}{8pt}
    \begin{tabular}{l|c|c|c|c|c}
        \toprule
        Method          & C-10  & C-100 & $\normalsize\tau$ & $\normalsize\rho$ & $\normalsize R$ \\\midrule
        FBNet v2 \cite{wan2020fbnetv2}       & 93.14     & 70.72  &-&-&-\\
        TuNAS  \cite{bender2020can}  & 92.78     & 70.11   &-&-&-\\
        CE \cite{hesslow2021contrastive} &90.55&70.78& 0.43 & 0.60 & 0.60 \\
        \textbf{BossNAS}  &  \textbf{93.29}  & \textbf{70.86}   &\textbf{0.59}  &\textbf{0.76}  &\textbf{0.79}\\
        \bottomrule
    \end{tabular}\vspace{-8pt}
    \caption{Comparison of searched model accuracy and architecture rating accuracy of different NAS methods on \textbf{NATS-Bench $\boldsymbol{\mathcal{S_S}}$} (C-10: \textbf{CIFAR-10}, C-100: \textbf{CIFAR-100}).
    }
    \label{tab:SS_acc}\vspace{-15pt}
\end{table}

\vspace{-5pt}
\subsection{Results on NATS-Bench $\boldsymbol{\mathcal{S_S}}$}
\vspace{-5pt}
For NATS-Bench \textit{size} search space $\mathcal{S_S}$, experiments are conducted on two datasets: CIFAR-10 and CIFAR-100.
Candidates of different channel numbers in our supernet share the weights in a slimmable manner \cite{Yu2019SlimmableNN,Yu2019UniversallySN,yu2019autoslim,LiWWLLC21,Chen2021AutoFormerST}.

\noindent\textbf{Performance of searched models.}
After searching on our supernet, we look up the performance of searched models in NATS-Bench $\mathcal{S_S}$ for fair comparision.
The results are shown in Tab. \ref{tab:SS_acc}. Our BossNAS outperforms recent NAS methods \cite{wan2020fbnetv2, bender2020can} designed particularly for network size search spaces, proving the generalization ability of our method on specified search spaces and relatively small datasets.

\noindent\textbf{Architecture rating accuracy.}
We rate all the 32768 architectures in the search space to compare with their ground truth accuracies in the benchmark on CIFAR-10 dataset. As shown in Fig. \ref{fig:MB_rank} right, all the architectures in the search space forms a dense, spindle-shaped pattern, proving the effectiveness of our BossNAS.

In addition, the architecture rating accuracies on CIFAR-100 dataset are shown in Tab. \ref{tab:SS_acc}. Our method, without access to the ground truth architecture accuracies and even without access to \textit{any} human-annotated labels, outperforms a predictor-based NAS method \cite{hesslow2021contrastive}, which is trained with ground truth architecture accuracies, by a large gap (\textit{i.e.} \textbf{0.16 $\normalsize\boldsymbol\tau$} and \textbf{0.19 $\normalsize\boldsymbol R$}).
More analysis on NATS-Bench $\mathcal{S_S}$ could be found in Appendix \ref{sec:app_nats}.
\begin{table}
\vspace{-8pt}
    \footnotesize
    \centering
    \begin{tabular}{l|l|c|c|c}
        \toprule
        Training & Evaluation                                                 &$\normalsize\tau$         & $\normalsize\rho$        & $\normalsize R$ \\
        \midrule
        $\mathtt{Supv.}$   distill.   & $\mathtt{Supv.}$   distill.           &0.62           &0.77           &0.83\\
        $\mathtt{Supv.}$   class. & $\mathtt{Supv.}$   class.         &0.46           &0.65           &0.71\\
        $\mathtt{Unsupv.}$ bootstrap.  & {\tt\textbf{Unsupv.}} \textbf{eval}        &  0.12        & 0.15           & 0.28 \\
        {\tt\textbf{Unsupv.}} \textbf{EB}    & $\mathtt{Supv.}$ linear eval    & 0.55          &0.73             &0.79\\
        {\tt\textbf{Unsupv.}} \textbf{EB}    & {\tt\textbf{Unsupv.}} \textbf{eval}    &\textbf{0.65}  &\textbf{0.78}  &\textbf{0.85}\\
        \bottomrule
    \end{tabular}
    \vspace{-7pt}
    \caption{Ablation analysis of training methods and evaluation methods on \textbf{MBConv Search Space}.
    }
    \label{tab:abl_training}
    \vspace{-14pt}
\end{table}
\vspace{-6pt}
\subsection{Ablation Study}
\vspace{-6pt}
In this section, we perform extensive ablation studies on MBConv search space and ImageNet to analyze our proposed training and evaluation methods separately. 

\noindent\textbf{training methods.}
We compared several training methods for the block-wise supernet: \textbf{(1)} \textit{Supervised distillation} method ($\mathtt{Supv.}$ distill.), using a pre-trained teacher model to provide block-wise supervision, \textit{i.e.} the training scheme used in DNA \cite{Li2020Blockwisely_cvpr} \textbf{(2)} \textit{Supervised classification} ($\mathtt{Supv.}$   class.), using real labels directly as the block-wise supervision. \textbf{(3)} \textit{Unsupervised bootstrapping} ($\mathtt{Unsupv.}$ bootstrap.), where the Siamese supernets are optimized by bootstrapping the corresponding paths in the two networks. \textbf{(4)} Our \textit{unsupervised ensemble bootstrapping} method ($\mathtt{Unsupv.}$ EB), where each sampled paths are optimized by learning to predict the probability ensemble of sampled paths from the mean teacher. As shown in Tab. \ref{tab:abl_training}, our training method surpasses all others, achieving the best results in architecture rating accuracy. In particular, by comparing the 3-\textit{rd} and 5-\textit{th} line, we can see that replacing our proposed $\mathtt{Unsupv.}$ EB with the naive $\mathtt{Unsupv.}$ bootstrap. scheme, the architecture rating accuracy drops sharply by \textbf{0.53} $\normalsize\boldsymbol\tau$. Without the probability ensemble, bootstrapping fails to reach a reasonable rating accuracy, proving that the proposed ensemble bootstrapping is indispensable for our BossNAS.

\noindent\textbf{Evaluation methods.}
Slimilar to the ablation analysis of training methods, we also compare our evaluation methods with \textbf{(1)} \textit{Supervised distillation} method ($\mathtt{Supv.}$ distill.) and \textbf{(2)} \textit{Supervised classification} ($\mathtt{Supv.}$   class.). Additionally, to perform ablation analysis of evaluation without changing the training method, we also compare with \textbf{(3)} supervised linear evaluation ($\mathtt{Supv.}$ linear eval), where architectures are rated by fixing the weights of the supernet and finetuning a weight sharing linear classifier to evaluate each architecture. \textbf{(4)} Our unsupervised evaluation metric ($\mathtt{Unsupv.}$ eval) rate architectures by its distance to the ensemble probability center of the whole search space. From the last two rows of Tab. \ref{tab:abl_training}, we suprisingly found that our $\mathtt{Unsupv.}$ eval outperforms supervised linear evaluation scheme in architecture rating by a remarkable gap (\textbf{0.1} $\normalsize\boldsymbol\tau$).%
\vspace{-6pt}
\subsection{Convergence Behavior}
\begin{figure}[t]
\vspace{-10pt}
    \centering
    \includegraphics[width=\linewidth]{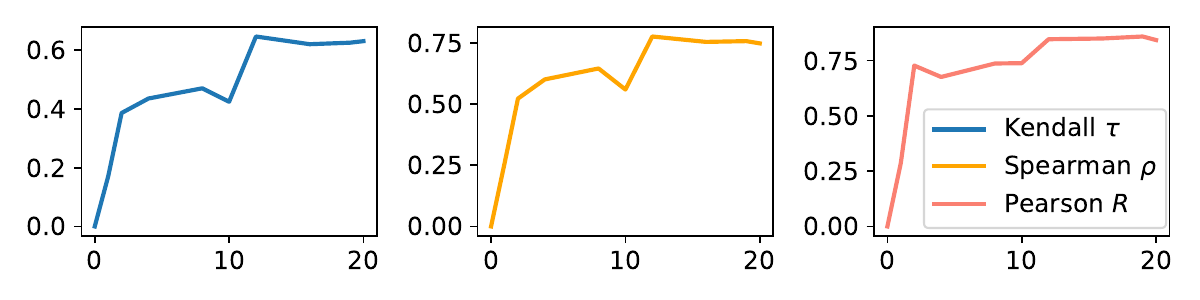}
    \vspace{-25pt}
    \caption{Ranking correlations during supernet training.}
    \label{fig:convergence}
    \vspace{-13pt}
\end{figure}

\vspace{-6pt}
To further demonstrate the effectiveness of BossNAS, we investigate the architecture rating accuracy during the supernet training process on MBConv search space with ImageNet. The three ranking correlation metrics of our BossNAS during its 20 training epochs are shown in Fig. \ref{fig:convergence}. The architecture rating accuracy increases rapidly in the early stage and continues to grow with minor fluctuation. The rating accuracy converges at the 12-th epoch and continues to be stable till the end of the training phase. The stably increasing architecture rating ability proves the stability of our BossNAS. In addition, the fast converging ranking correlation demonstrates that our method is easy to optimize and do not require longer training. Please refer to Appendix \ref{sec:app_nats} for analysis of convergence behavior on NATS-Bench $\mathcal{S_S}$.

\vspace{-9pt}
\section{Conclusion}
\vspace{-7pt}
In this work, we present BossNAS, a general, unsupervised NAS method with the \textit{ensemble bootstrapping} training technique and an \textit{unsupervised evaluation metric}. Experiments on three search spaces prove that our method successfully addressed the problem of inaccurate architecture rating caused by large weight-sharing space while avoiding the architectural bias brought by supervised distillation. Ablation analysis proved that the two components, \textit{ensemble bootstrapping} scheme and \textit{unsupervised evaluation metric}, are both crucial for our method. Additionally, we present a fabric-like search space named HyTra. On this challenging search space, our searched hybrid CNN-transformer model, achieves 82.5\% accuracy on ImageNet, surpassing EfficientNet by 2.4\% with comparable compute time.

\vspace{-9pt}
\section*{Acknowledgement}
\vspace{-7pt}

This work was supported in part by National Key R\&D Program of China under Grant No. 2020AAA0109700 and the funding of ``Leading Innovation Team of the Zhejiang Province" (2018R01017). Dr Xiaojun Chang is partially supported by Australian Research Council (ARC) Discovery Early Career Research Award (DECRA) under grant no. DE190100626 and Intelligence Advanced Research Projects Activity (IARPA) via Department of Interior/Interior Business Center (DOI/IBC).

{\small
\bibliographystyle{ieee_fullname}
\bibliography{egbib}

\begin{thebibliography}{10}\itemsep=-1pt

\bibitem{akimoto2019adaptive}
Youhei Akimoto, Shinichi Shirakawa, Nozomu Yoshinari, Kento Uchida, Shota
  Saito, and Kouhei Nishida.
\newblock Adaptive stochastic natural gradient method for one-shot neural
  architecture search.
\newblock In {\em {ICML}}, 2019.

\bibitem{baker2016designing}
Bowen Baker, Otkrist Gupta, Nikhil Naik, and Ramesh Raskar.
\newblock Designing neural network architectures using reinforcement learning.
\newblock In {\em {ICLR}}, 2017.

\bibitem{bello2021lambdanetworks}
Irwan Bello.
\newblock Lambdanetworks: Modeling long-range interactions without attention.
\newblock In {\em {ICLR}}, 2021.

\bibitem{bender2018understanding}
Gabriel Bender, Pieter{-}Jan Kindermans, Barret Zoph, Vijay Vasudevan, and
  Quoc~V. Le.
\newblock Understanding and simplifying one-shot architecture search.
\newblock In {\em {ICML}}, 2018.

\bibitem{bender2020can}
Gabriel Bender, Hanxiao Liu, Bo Chen, Grace Chu, Shuyang Cheng, Pieter-Jan
  Kindermans, and Quoc~V Le.
\newblock Can weight sharing outperform random architecture search? an
  investigation with tunas.
\newblock In {\em {CVPR}}, 2020.

\bibitem{brock2017smash}
Andrew Brock, Theodore Lim, James~M. Ritchie, and Nick Weston.
\newblock {SMASH:} one-shot model architecture search through hypernetworks.
\newblock In {\em {ICLR}}, 2018.

\bibitem{Cai2018ProxylessNASDN}
Han Cai, Ligeng Zhu, and Song Han.
\newblock Proxylessnas: Direct neural architecture search on target task and
  hardware.
\newblock In {\em {ICLR}}, 2019.

\bibitem{carion2020detr}
Nicolas Carion, Francisco Massa, Gabriel Synnaeve, Nicolas Usunier, Alexander
  Kirillov, and Sergey Zagoruyko.
\newblock End-to-end object detection with transformers.
\newblock In {\em {ECCV}}, 2020.

\bibitem{Chen2021AutoFormerST}
Minghao Chen, Houwen Peng, Jianlong Fu, and Haibin Ling.
\newblock {AutoFormer}: Searching transformers for visual recognition.
\newblock In {\em {ICCV}}, 2021.

\bibitem{chen2020simple}
Ting Chen, Simon Kornblith, Mohammad Norouzi, and Geoffrey Hinton.
\newblock A simple framework for contrastive learning of visual
  representations.
\newblock In {\em {ICML}}, 2020.

\bibitem{chen2020exploring}
Xinlei Chen and Kaiming He.
\newblock Exploring simple siamese representation learning.
\newblock In {\em {CVPR}}, 2021.

\bibitem{chen2018reinforced}
Yukang Chen, Gaofeng Meng, Qian Zhang, Shiming Xiang, Chang Huang, Lisen Mu,
  and Xinggang Wang.
\newblock {RENAS:} reinforced evolutionary neural architecture search.
\newblock In {\em {CVPR}}, 2019.

\bibitem{ChengZHDCLDG20}
Xuelian Cheng, Yiran Zhong, Mehrtash Harandi, Yuchao Dai, Xiaojun Chang,
  Hongdong Li, Tom Drummond, and Zongyuan Ge.
\newblock Hierarchical neural architecture search for deep stereo matching.
\newblock In {\em {NeurIPS}}, 2020.

\bibitem{chu2021conditional}
Xiangxiang Chu, Zhi Tian, Bo Zhang, Xinlong Wang, Xiaolin Wei, Huaxia Xia, and
  Chunhua Shen.
\newblock Conditional positional encodings for vision transformers.
\newblock {\em arXiv:2102.10882}, 2021.

\bibitem{chu2021fairnas}
Xiangxiang Chu, Bo Zhang, and Ruijun Xu.
\newblock Fair{NAS}: Rethinking evaluation fairness of weight sharing neural
  architecture search.
\newblock In {\em {ICCV}}, 2021.

\bibitem{deng2009imagenet}
Jia Deng, Wei Dong, Richard Socher, Li-Jia Li, Kai Li, and Li Fei-Fei.
\newblock Imagenet: A large-scale hierarchical image database.
\newblock In {\em {CVPR}}, 2009.

\bibitem{dong2021nats}
Xuanyi Dong, Lu Liu, Katarzyna Musial, and Bogdan Gabrys.
\newblock {NATS-Bench}: Benchmarking nas algorithms for architecture topology
  and size.
\newblock {\em IEEE Transactions on Pattern Analysis and Machine Intelligence
  (TPAMI)}, 2021.
\newblock \mbox{doi}:\url{10.1109/TPAMI.2021.3054824}.

\bibitem{dong2019searching}
Xuanyi Dong and Yi Yang.
\newblock Searching for a robust neural architecture in four {GPU} hours.
\newblock In {\em {CVPR}}, 2019.

\bibitem{dong2020nasbench201}
Xuanyi Dong and Yi Yang.
\newblock {NAS}-{B}ench-201: Extending the scope of reproducible neural
  architecture search.
\newblock In {\em {ICLR}}, 2020.

\bibitem{dosovitskiy2021vit}
Alexey Dosovitskiy, Lucas Beyer, Alexander Kolesnikov, Dirk Weissenborn,
  Xiaohua Zhai, Thomas Unterthiner, Mostafa Dehghani, Matthias Minderer, Georg
  Heigold, Sylvain Gelly, Jakob Uszkoreit, and Neil Houlsby.
\newblock An image is worth 16x16 words: Transformers for image recognition at
  scale.
\newblock In {\em {ICLR}}, 2021.

\bibitem{grill2020bootstrap}
Jean-Bastien Grill, Florian Strub, Florent Altch{\'e}, Corentin Tallec,
  Pierre~H Richemond, Elena Buchatskaya, Carl Doersch, Bernardo~Avila Pires,
  Zhaohan~Daniel Guo, Mohammad~Gheshlaghi Azar, et~al.
\newblock Bootstrap your own latent: A new approach to self-supervised
  learning.
\newblock In {\em {NeurIPS}}, 2020.

\bibitem{Guo2019SinglePO}
Zichao Guo, Xiangyu Zhang, Haoyuan Mu, Wen Heng, Zechun Liu, Yichen Wei, and
  Jian Sun.
\newblock Single path one-shot neural architecture search with uniform
  sampling.
\newblock In {\em ECCV}, 2020.

\bibitem{han2021tnt}
Kai Han, An Xiao, Enhua Wu, Jianyuan Guo, Chunjing Xu, and Yunhe Wang.
\newblock Transformer in transformer.
\newblock {\em arXiv:2103.00112}, 2021.

\bibitem{he2020momentum}
Kaiming He, Haoqi Fan, Yuxin Wu, Saining Xie, and Ross Girshick.
\newblock Momentum contrast for unsupervised visual representation learning.
\newblock In {\em {CVPR}}, 2020.

\bibitem{he2016resnet}
Kaiming He, Xiangyu Zhang, Shaoqing Ren, and Jian Sun.
\newblock Deep residual learning for image recognition.
\newblock In {\em {CVPR}}, 2016.

\bibitem{he2021transreid}
Shuting He, Hao Luo, Pichao Wang, Fan Wang, Hao Li, and Wei Jiang.
\newblock Trans{R}e{ID}: Transformer-based object re-identification.
\newblock In {\em {ICCV}}, 2021.

\bibitem{hesslow2021contrastive}
Daniel Hesslow and Iacopo Poli.
\newblock Contrastive embeddings for neural architectures.
\newblock {\em arXiv:2102.04208}, 2021.

\bibitem{hjelm2018learning}
R~Devon Hjelm, Alex Fedorov, Samuel Lavoie-Marchildon, Karan Grewal, Phil
  Bachman, Adam Trischler, and Yoshua Bengio.
\newblock Learning deep representations by mutual information estimation and
  maximization.
\newblock In {\em {ICLR}}, 2019.

\bibitem{Howard2019SearchingFM}
Andrew Howard, Mark Sandler, Grace Chu, Liang{-}Chieh Chen, Bo Chen, Mingxing
  Tan, Weijun Wang, Yukun Zhu, Ruoming Pang, Vijay Vasudevan, Quoc~V. Le, and
  Hartwig Adam.
\newblock Searching for {M}obile{N}et{V}3.
\newblock In {\em {ICCV}}, 2019.

\bibitem{howard2017mobilenets}
Andrew~G Howard, Menglong Zhu, Bo Chen, Dmitry Kalenichenko, Weijun Wang,
  Tobias Weyand, Marco Andreetto, and Hartwig Adam.
\newblock Mobilenets: Efficient convolutional neural networks for mobile vision
  applications.
\newblock {\em arXiv:1704.04861}, 2017.

\bibitem{Hu2019SqueezeandExcitationN}
Jie Hu, Li Shen, and Gang Sun.
\newblock Squeeze-and-excitation networks.
\newblock In {\em {CVPR}}, 2018.

\bibitem{hu2020angle}
Yiming Hu, Yuding Liang, Zichao Guo, Ruosi Wan, Xiangyu Zhang, Yichen Wei,
  Qingyi Gu, and Jian Sun.
\newblock Angle-based search space shrinking for neural architecture search.
\newblock In {\em {ECCV}}, 2020.

\bibitem{jiang2021transgan}
Yifan Jiang, Shiyu Chang, and Zhangyang Wang.
\newblock Transgan: Two transformers can make one strong gan.
\newblock {\em arXiv:2102.07074}, 2021.

\bibitem{kendall1938new}
Maurice~G Kendall.
\newblock A new measure of rank correlation.
\newblock {\em Biometrika}, 30(1/2):81--93, 1938.

\bibitem{komodakis2018unsupervised}
Nikos Komodakis and Spyros Gidaris.
\newblock Unsupervised representation learning by predicting image rotations.
\newblock In {\em {ICLR}}, 2018.

\bibitem{Krizhevsky09cifar}
A. Krizhevsky and G. Hinton.
\newblock Learning multiple layers of features from tiny images.
\newblock {\em Master's thesis, Department of Computer Science, University of
  Toronto}, 2009.

\bibitem{Li2020Blockwisely_cvpr}
Changlin Li, Jiefeng Peng, Liuchun Yuan, Guangrun Wang, Xiaodan Liang, Liang
  Lin, and Xiaojun Chang.
\newblock Blockwisely supervised neural architecture search with knowledge
  distillation.
\newblock In {\em {CVPR}}, 2020.

\bibitem{LiWWLLC21}
Changlin Li, Guangrun Wang, Bing Wang, Xiaodan Liang, Zhihui Li, and Xiaojun
  Chang.
\newblock {Dynamic Slimmable Network}.
\newblock In {\em {CVPR}}, 2021.

\bibitem{li2020improving}
Xiang Li, Chen Lin, Chuming Li, Ming Sun, Wei Wu, Junjie Yan, and Wanli Ouyang.
\newblock Improving one-shot nas by suppressing the posterior fading.
\newblock In {\em {CVPR}}, 2020.

\bibitem{liu2019auto}
Chenxi Liu, Liang-Chieh Chen, Florian Schroff, Hartwig Adam, Wei Hua, Alan~L
  Yuille, and Li Fei-Fei.
\newblock Auto-deeplab: Hierarchical neural architecture search for semantic
  image segmentation.
\newblock In {\em {CVPR}}, 2019.

\bibitem{liu2020unnas}
Chenxi Liu, Piotr Doll{\'a}r, Kaiming He, Ross Girshick, Alan Yuille, and
  Saining Xie.
\newblock Are labels necessary for neural architecture search?
\newblock In {\em {ECCV}}, 2020.

\bibitem{liu2018progressive}
Chenxi Liu, Barret Zoph, Maxim Neumann, Jonathon Shlens, Wei Hua, Li-Jia Li, Li
  Fei-Fei, Alan Yuille, Jonathan Huang, and Kevin Murphy.
\newblock Progressive neural architecture search.
\newblock In {\em {ECCV}}, 2018.

\bibitem{Liu2018DARTSDA}
Hanxiao Liu, Karen Simonyan, and Yiming Yang.
\newblock {DARTS:} differentiable architecture search.
\newblock In {\em {ICLR}}, 2019.

\bibitem{loshchilov2016sgdr}
Ilya Loshchilov and Frank Hutter.
\newblock Sgdr: Stochastic gradient descent with warm restarts.
\newblock In {\em {ICLR}}, 2017.

\bibitem{maennel2020neural}
Hartmut Maennel, Ibrahim~M Alabdulmohsin, Ilya~O Tolstikhin, Robert Baldock,
  Olivier Bousquet, Sylvain Gelly, and Daniel Keysers.
\newblock What do neural networks learn when trained with random labels?
\newblock In {\em {NeurIPS}}, 2020.

\bibitem{moons2020donna}
Bert Moons, Parham Noorzad, Andrii Skliar, Giovanni Mariani, Dushyant Mehta,
  Chris Lott, and Tijmen Blankevoort.
\newblock Distilling optimal neural networks: Rapid search in diverse spaces.
\newblock {\em arXiv:2012.08859}, 2020.

\bibitem{negrinho2017deeparchitect}
Renato Negrinho and Geoffrey~J. Gordon.
\newblock Deeparchitect: Automatically designing and training deep
  architectures.
\newblock {\em arXiv:1704.08792}, 2017.

\bibitem{noroozi2016unsupervised}
Mehdi Noroozi and Paolo Favaro.
\newblock Unsupervised learning of visual representations by solving jigsaw
  puzzles.
\newblock In {\em {ECCV}}, 2016.

\bibitem{oord2018representation}
Aaron van~den Oord, Yazhe Li, and Oriol Vinyals.
\newblock Representation learning with contrastive predictive coding.
\newblock {\em arXiv:1807.03748}, 2018.

\bibitem{parmar2018imagetransformer}
Niki Parmar, Ashish Vaswani, Jakob Uszkoreit, Lukasz Kaiser, Noam Shazeer,
  Alexander Ku, and Dustin Tran.
\newblock Image transformer.
\newblock In {\em ICML}, 2018.

\bibitem{Peng2021PiNAS}
Jiefeng Peng, Jiqi Zhang, Changlin Li, Guangrun Wang, Xiaodan Liang, and Liang
  Lin.
\newblock {Pi-NAS}: Improving neural architecture search by reducing supernet
  training consistency shift.
\newblock In {\em {ICCV}}, 2021.

\bibitem{pham2018enas}
Hieu Pham, Melody Guan, Barret Zoph, Quoc Le, and Jeff Dean.
\newblock Efficient neural architecture search via parameters sharing.
\newblock In {\em {ICML}}, 2018.

\bibitem{Pham2018EfficientNA}
Hieu Pham, Melody~Y. Guan, Barret Zoph, Quoc~V. Le, and Jeff Dean.
\newblock Efficient neural architecture search via parameter sharing.
\newblock In {\em {ICML}}, 2018.

\bibitem{Real2019AmoebaNet}
Esteban Real, Alok Aggarwal, Yanping Huang, and Quoc~V. Le.
\newblock Regularized evolution for image classifier architecture search.
\newblock In {\em {AAAI}}, 2019.

\bibitem{RenXCHLCW2020}
Pengzhen Ren, Yun Xiao, Xiaojun Chang, Po-Yao Huang, Zhihui Li, Xiaojiang Chen,
  and Xin Wang.
\newblock A comprehensive survey of neural architecture search: Challenges and
  solutions.
\newblock {\em {ACM Computing Surveys}}, 2021.

\bibitem{Sandler2018MobileNetV2IR}
Mark Sandler, Andrew~G. Howard, Menglong Zhu, Andrey Zhmoginov, and
  Liang{-}Chieh Chen.
\newblock Mobilenetv2: Inverted residuals and linear bottlenecks.
\newblock In {\em {CVPR}}, 2018.

\bibitem{saxena2016convolutional}
Shreyas Saxena and Jakob Verbeek.
\newblock Convolutional neural fabrics.
\newblock In {\em {NeurIPS}}, 2016.

\bibitem{sciuto2019evaluating}
Christian Sciuto, Kaicheng Yu, Martin Jaggi, Claudiu Musat, and Mathieu
  Salzmann.
\newblock Evaluating the search phase of neural architecture search.
\newblock In {\em {ICLR}}, 2020.

\bibitem{Meal}
Zhiqiang Shen, Zhankui He, and Xiangyang Xue.
\newblock Meal: Multi-model ensemble via adversarial learning.
\newblock In {\em {AAAI}}, 2019.

\bibitem{Mealv2}
Zhiqiang Shen and Marios Savvides.
\newblock Meal v2: Boosting vanilla resnet-50 to 80\%+ top-1 accuracy on
  imagenet without tricks.
\newblock {\em arXiv:2009.08453}, 2020.

\bibitem{shi2020bridging}
Han Shi, Renjie Pi, Hang Xu, Zhenguo Li, James Kwok, and Tong Zhang.
\newblock Bridging the gap between sample-based and one-shot neural
  architecture search with bonas.
\newblock In {\em {NeurIPS}}, 2020.

\bibitem{srinivas2021bottleneck}
Aravind Srinivas, Tsung-Yi Lin, Niki Parmar, Jonathon Shlens, Pieter Abbeel,
  and Ashish Vaswani.
\newblock Bottleneck transformers for visual recognition.
\newblock In {\em {CVPR}}, 2021.

\bibitem{Tan2018MnasNetPN}
Mingxing Tan, Bo Chen, Ruoming Pang, Vijay Vasudevan, Mark Sandler, Andrew
  Howard, and Quoc~V. Le.
\newblock Mnasnet: Platform-aware neural architecture search for mobile.
\newblock In {\em {CVPR}}, 2019.

\bibitem{Tan2019EfficientNetRM}
Mingxing Tan and Quoc~V. Le.
\newblock Efficientnet: Rethinking model scaling for convolutional neural
  networks.
\newblock In {\em {ICML}}, 2019.

\bibitem{tian2019contrastive}
Yonglong Tian, Dilip Krishnan, and Phillip Isola.
\newblock Contrastive multiview coding.
\newblock {\em arXiv:1906.05849}, 2019.

\bibitem{touvron2020deit}
Hugo Touvron, Matthieu Cord, Matthijs Douze, Francisco Massa, Alexandre
  Sablayrolles, and Herv{\'e} J{\'e}gou.
\newblock Training data-efficient image transformers \& distillation through
  attention.
\newblock In {\em {ICML}}, 2021.

\bibitem{wan2020fbnetv2}
Alvin Wan, Xiaoliang Dai, Peizhao Zhang, Zijian He, Yuandong Tian, Saining Xie,
  Bichen Wu, Matthew Yu, Tao Xu, Kan Chen, et~al.
\newblock Fbnetv2: Differentiable neural architecture search for spatial and
  channel dimensions.
\newblock In {\em {CVPR}}, 2020.

\bibitem{wang2018non}
Xiaolong Wang, Ross Girshick, Abhinav Gupta, and Kaiming He.
\newblock Non-local neural networks.
\newblock In {\em {CVPR}}, 2018.

\bibitem{wei2020self}
Chen Wei, Yiping Tang, Chuang Niu, Haihong Hu, Yue Wang, and Jimin Liang.
\newblock Self-supervised representation learning for evolutionary neural
  architecture search.
\newblock {\em arXiv:2011.00186}, 2020.

\bibitem{Wu2018FBNetHE}
Bichen Wu, Xiaoliang Dai, Peizhao Zhang, Yanghan Wang, Fei Sun, Yiming Wu,
  Yuandong Tian, Peter Vajda, Yangqing Jia, and Kurt Keutzer.
\newblock Fbnet: Hardware-aware efficient convnet design via differentiable
  neural architecture search.
\newblock In {\em {CVPR}}, 2019.

\bibitem{wu2018unsupervised}
Zhirong Wu, Yuanjun Xiong, Stella~X Yu, and Dahua Lin.
\newblock Unsupervised feature learning via non-parametric instance
  discrimination.
\newblock In {\em {CVPR}}, 2018.

\bibitem{yan2020does}
Shen Yan, Yu Zheng, Wei Ao, Xiao Zeng, and Mi Zhang.
\newblock Does unsupervised architecture representation learning help neural
  architecture search?
\newblock In {\em NeurIPS}, 2020.

\bibitem{anonymous2020nas}
Antoine Yang, Pedro~M. Esperan{\c{c}}a, and Fabio~Maria Carlucci.
\newblock {NAS} evaluation is frustratingly hard.
\newblock In {\em {ICLR}}, 2020.

\bibitem{ying2019bench101}
Chris Ying, Aaron Klein, Eric Christiansen, Esteban Real, Kevin Murphy, and
  Frank Hutter.
\newblock Nas-bench-101: Towards reproducible neural architecture search.
\newblock In {\em {ICML}}, 2019.

\bibitem{you2017scaling}
Yang You, Igor Gitman, and Boris Ginsburg.
\newblock Scaling sgd batch size to 32k for imagenet training.
\newblock {\em arXiv:1708.03888}, 2017.

\bibitem{yu2019autoslim}
Jiahui Yu and Thomas Huang.
\newblock Autoslim: Towards one-shot architecture search for channel numbers.
\newblock {\em arXiv:1903.11728}, 2019.

\bibitem{Yu2019UniversallySN}
Jiahui Yu and Thomas~S. Huang.
\newblock Universally slimmable networks and improved training techniques.
\newblock In {\em {ICCV}}, 2019.

\bibitem{Yu2019SlimmableNN}
Jiahui Yu, Linjie Yang, Ning Xu, Jianchao Yang, and Thomas~S. Huang.
\newblock Slimmable neural networks.
\newblock In {\em {ICLR}}, 2019.

\bibitem{yuan2021t2t}
Li Yuan, Yunpeng Chen, Tao Wang, Weihao Yu, Yujun Shi, Francis~EH Tay, Jiashi
  Feng, and Shuicheng Yan.
\newblock Tokens-to-token vit: Training vision transformers from scratch on
  imagenet.
\newblock In {\em {ICCV}}, 2021.

\bibitem{zela2019bench1s1}
Arber Zela, Julien Siems, and Frank Hutter.
\newblock Nas-bench-1shot1: Benchmarking and dissecting one-shot neural
  architecture search.
\newblock In {\em {ICLR}}, 2019.

\bibitem{zhang2016understanding}
Chiyuan Zhang, Samy Bengio, Moritz Hardt, Benjamin Recht, and Oriol Vinyals.
\newblock Understanding deep learning requires rethinking generalization.
\newblock {\em arXiv:1611.03530}, 2016.

\bibitem{ZhangLPCGS20}
Miao Zhang, Huiqi Li, Shirui Pan, Xiaojun Chang, Zongyuan Ge, and Steven~W. Su.
\newblock Differentiable neural architecture search in equivalent space with
  exploration enhancement.
\newblock In {\em {NeurIPS}}, 2020.

\bibitem{ZhangLPCS20}
Miao Zhang, Huiqi Li, Shirui Pan, Xiaojun Chang, and Steven~W. Su.
\newblock Overcoming multi-model forgetting in one-shot {NAS} with diversity
  maximization.
\newblock In {\em {CVPR}}, 2020.

\bibitem{zhang2020stagewisepruning}
Mingyang Zhang and Linlin Ou.
\newblock Stage-wise channel pruning for model compression.
\newblock {\em arXiv:2011.04908}, 2020.

\bibitem{zhang2021semidnagan}
Man Zhang, Yong Zhou, Jiaqi Zhao, Shixiong Xia, Jiaqi Wang, and Zizheng Huang.
\newblock Semi-supervised blockwisely architecture search for efficient
  lightweight generative adversarial network.
\newblock {\em Pattern Recognition}, 112:107794, 2021.

\bibitem{zhang2016colorful}
Richard Zhang, Phillip Isola, and Alexei~A Efros.
\newblock Colorful image colorization.
\newblock In {\em {ECCV}}, 2016.

\bibitem{zhang2021randomlabelnas}
Xuanyang Zhang, Pengfei Hou, Xiangyu Zhang, and Jian Sun.
\newblock Neural architecture search with random labels.
\newblock In {\em {CVPR}}, 2021.

\bibitem{zheng2020setr}
Sixiao Zheng, Jiachen Lu, Hengshuang Zhao, Xiatian Zhu, Zekun Luo, Yabiao Wang,
  Yanwei Fu, Jianfeng Feng, Tao Xiang, Philip~HS Torr, et~al.
\newblock Rethinking semantic segmentation from a sequence-to-sequence
  perspective with transformers.
\newblock In {\em {CVPR}}, 2021.

\bibitem{zhong2018practical}
Zhao Zhong, Junjie Yan, Wei Wu, Jing Shao, and Cheng{-}Lin Liu.
\newblock Practical block-wise neural network architecture generation.
\newblock In {\em {CVPR}}, 2018.

\bibitem{zhu2021deformabledetr}
Xizhou Zhu, Weijie Su, Lewei Lu, Bin Li, Xiaogang Wang, and Jifeng Dai.
\newblock Deformable {DETR}: Deformable transformers for end-to-end object
  detection.
\newblock In {\em {ICLR}}, 2021.

\bibitem{zhuang2019local}
Chengxu Zhuang, Alex~Lin Zhai, and Daniel Yamins.
\newblock Local aggregation for unsupervised learning of visual embeddings.
\newblock In {\em {ICCV}}, 2019.

\bibitem{zoph2016neural}
Barret Zoph and Quoc~V. Le.
\newblock Neural architecture search with reinforcement learning.
\newblock In {\em {ICLR}}, 2017.

\bibitem{zoph2018learning}
Barret Zoph, Vijay Vasudevan, Jonathon Shlens, and Quoc~V Le.
\newblock Learning transferable architectures for scalable image recognition.
\newblock In {\em {CVPR}}, 2018.

\end{thebibliography}
}

\renewcommand{\appendixname}{Appendix~\Alph{section}}
\clearpage
\emptythanks
\setcounter{footnote}{0}

\appendix

\title{BossNAS: Exploring Hybrid CNN-transformers with Block-wisely Self-supervised Neural Architecture Search\\[7pt]\large{Supplementary Material}}
\author{Changlin Li$^{1}$, \quad Tao Tang$^2$, \quad Guangrun Wang$^{3,4}$, \quad Jiefeng Peng$^{3}$, \quad Bing Wang$^{5}$,\\ \quad Xiaodan Liang$^{2}$\thanks{Corresponding Author.}, \quad Xiaojun Chang$^{6}$\\
\small$^1$GORSE Lab, Dept. of DSAI,
Monash University \quad \small$^2$Sun Yat-sen University\quad \small$^3$DarkMatter AI Research \\ \quad \small$^4$University of Oxford \quad \small$^5$Alibaba Group \quad \small $^6$RMIT University\\
{\tt\small changlin.li@monash.edu, }\\{\tt\small \{trent.tangtao,wanggrun,jiefengpeng,xdliang328\}@gmail.com,}\\{\tt\small fengquan.wb@alibaba-inc.com, xiaojun.chang@rmit.edu.au}
}
\maketitle
\section{Appendix}
\subsection{A brief review of NAS}
NAS methods aim to automatically optimize neural network architectures by exploring search spaces with \textit{search algorithms} and evaluating architectures by means of \textit{rating schemes}. NAS methods can be divided into two categories depending on the rating scheme utilized, \textit{i.e.} multi-trial NAS and weight-sharing NAS. \textbf{Multi-trial NAS} methods \cite{zoph2016neural,baker2016designing,Real2019AmoebaNet,Tan2018MnasNetPN,liu2018progressive,ZhangLPCS20} rate all sampled architectures by training them from scratch, making this process computationally prohibitive and difficult to deploy on large datasets.
They either perform architecture rating by training on relatively small datasets (\textit{e.g.} CIFAR-10) \cite{zoph2016neural,baker2016designing,Real2019AmoebaNet} or by training for the first few epochs (\textit{e.g.} 5 epochs) \cite{Tan2018MnasNetPN} on ImageNet.
To avoid repeated training of candidate networks, \textbf{weight-sharing NAS} methods \cite{Cai2018ProxylessNASDN, Liu2018DARTSDA, dong2019searching, akimoto2019adaptive, brock2017smash,ChengZHDCLDG20,ZhangLPCGS20} optimize a \textit{supernet} that encodes the whole search space, then rate each candidate architecture according to its weights inherited from the supernet. Among them, \textit{gradient-based} approaches \cite{Liu2018DARTSDA,Cai2018ProxylessNASDN, Wu2018FBNetHE} and \textit{sampler-based} approaches \cite{pham2018enas,shi2020bridging} jointly optimize the weight of the supernet and the factors (or agent) used to choose the architecture; for their part, \textit{one-shot} approaches\footnote{In this paper, following the pioneering works SMASH \cite{brock2017smash} and One-shot \cite{bender2018understanding}, when we refer to one-shot NAS methods, we are discussing those incorporating two-stage (i.e., a supernet training stage and a searching stage) weight-sharing methods rather than the general weight-sharing NAS discussed in \cite{zela2019bench1s1}.} \cite{Guo2019SinglePO, chu2021fairnas, brock2017smash, bender2018understanding,Peng2021PiNAS} optimize the supernet before performing a search with the frozen supernet weights. We refer to \cite{RenXCHLCW2020} for a more comprehensive NAS review.

\subsection{Implementation Details}\label{sec:app_details}
\noindent\textbf{Search spaces.} We evaluate our method on three search spaces:
\begin{itemize}
    \item \textbf{HyTra search space.} The beginning of the networks in this search space is the classic ResNet stem that reduces the spatial resolution by a factor of 4 with a strided $7$$\times$$7$ convolution layer and a max-pooling layer. It contains $L=16$ choice block layers in total, as the same to ResNet50. Before the first choice block layer, the input can be further down-sampled to different scales. The downsampling module consists of multiple $3$$\times$$3$ convolutions with stride of 2. At each choice block layer, the spatial resolution can either stay unchanged or be reduced to half of its scale, unless reaching the smallest scale $1/32$. As introduced in Sec. \ref{sec:searchspaceT}, this search space contains two disparate candidate choices: $\{\mathtt{ResConv, ResAtt}\}$. As transformer blocks are expensive in the first scales, we only enable the choice of {\tt \textbf{ResAtt}} in the last two scales (\textit{i.e.} $1/16$ and $1/32$). The total size of this challenging hybrid search space is roughly $2.8$$\times$$10^{6}$.
    \item \textbf{MBConv search space.} MobileNet-like search space and its variations are generally used as benchmarks for recent NAS methods \cite{Tan2018MnasNetPN,Howard2019SearchingFM,Tan2019EfficientNetRM,Cai2018ProxylessNASDN,Wu2018FBNetHE,chu2021fairnas,Li2020Blockwisely_cvpr,moons2020donna,zhang2021randomlabelnas}. Following Li \textit{et. al.} \cite{Li2020Blockwisely_cvpr}, we use a search space with 18 layers and each layer contains 4 candidate MobileNet blocks (combination of kernel size $\{3, 5\}$ and reduction rate $\{3, 6\}$). This results in a large search space containing about $4^{18}\approx6.9$$\times$$10^{10}$ architectures.
    \item \textbf{NATS-Bench $\boldsymbol{\mathcal{S_S}}$.} The NATS-Bench \textit{size} search space $\mathcal{S_S}$ \cite{dong2021nats} is a channel configuration search space built upon a fixed cell-based architecture with 5 layers, where the 2-nd and 4-th layers have a down-sample rate of 2. Number of channels in each layer is chosen from  \{8, 16, 24, 32, 40, 48, 56, 64\}. $\mathcal{S_S}$ has $8^5=32768$ architecture candidates in total.
     Candidates of different channel numbers in our supernet share the weights in a slimmable manner \cite{Yu2019SlimmableNN,Yu2019UniversallySN,yu2019autoslim,LiWWLLC21,Chen2021AutoFormerST}. We divide the supernet into 3 blocks, according to spatial size. 
\end{itemize}

\noindent\textbf{Datasets.}
The datasets we use to evaluate and analyze our method include ImageNet \cite{deng2009imagenet}, CIFAR-10 and CIFAR-100 \cite{Krizhevsky09cifar}. \textbf{ImageNet} is a large-scale dataset containing 1.2 M $\mathtt{train}$ set images and 50 K $\mathtt{val}$ set images in 1000 classes. We randomly samples 50 K images from the original $\mathtt{train}$ set to form a $\mathtt{NAS}$-$\mathtt{val}$ set for architecture rating and use the remainder as the $\mathtt{NAS}$-$\mathtt{train}$ set for supernet training. No labels are used during training and searching of our NAS method. Finally, our searched architectures are retrained from scratch on $\mathtt{train}$ set and evaluated on $\mathtt{val}$ set.
For \textbf{CIFAR-10} and \textbf{CIFAR-100} \cite{Krizhevsky09cifar}, we use the splits proposed in NATS-Bench \cite{dong2021nats}. CIFAR-10 is divided into 25 K $\mathtt{train}$ set, 25 K $\mathtt{val}$ set, and 10 K $\mathtt{test}$ set. CIFAR-100 is devided into 50 K $\mathtt{train}$ set, 5 K $\mathtt{val}$ set, and 5 K $\mathtt{test}$ set. The final accuracies of searched architectures are queried from NATS-Bench $\mathcal{S_S}$ \cite{dong2021nats}.\\[5pt]
%
\noindent\textbf{Training details.}

We train each block of the \textbf{BossNAS supernet} for 20 epochs including 1 linear warm-up epoch on ImageNet. For the relatively smaller CIFAR datasets, we extend it to 30 epochs. In each training step, we randomly sample 4 paths for the ensemble bootstrapping. Other hyperparameters for self-supervised training of the supernet follow closely to BYOL \cite{grill2020bootstrap}, we use the $\mathtt{LARS}$ optimizer \cite{you2017scaling} with a cosine decay learning rate schedule \cite{loshchilov2016sgdr}. The base learning rate is set to 4.8 for a total batchsize of 4096.

For ImageNet retraining of \textbf{BossNet-T models}, we follow similar with DeiT \cite{touvron2020deit}, as we found it robust for both CNNs and transformers. More specifically, we use AdamW optimizer with 1e-3 initial learning rate and cosine learning rate scheduler, for a total batch size of 1024. Weight decay is set to 0.05. We use model EMA with decay rate 0.99996 following \cite{yuan2021t2t}. Please refer to DeiT \cite{touvron2020deit} for more details on data-augmentation and regularization.

For ImageNet retraining of \textbf{BossNet-M models}, we follow closely to EfficientNet \cite{Tan2019EfficientNetRM}. We use batchsize 4096, RMSprop optimizer with momentum 0.9 and initial learning rate of 0.256 which decays by 0.97 every 2.4 epochs. Please refer to EfficientNet \cite{Tan2019EfficientNetRM} for more details of other settings.\\[5pt]
%
\noindent\textbf{Re-implementation of other NAS methods on HyTra.}

For DNA \cite{Li2020Blockwisely_cvpr}, we use ResNet-50 \cite{he2016resnet} as the teacher model. We divide the supernet into four blocks, with four layers in each block, and train each block for 20 epochs. The intermediate features of every block of the student supernet and the teacher are all downsampled with global pooling and projected with one fully-connected layer before calculating distillation loss, as the scale of different candidate block is not the same in HyTra search space. Other settings follow closely to DNA \cite{Li2020Blockwisely_cvpr}.

For UnNAS \cite{liu2020unnas}, we adopt \textit{rotation prediction} \cite{komodakis2018unsupervised} ($\mathtt{Rot}$) pretext task, for its simplicity. Following \cite{liu2020unnas}, we use three extra stride-2 convolution layers at the beginning of the supernet to reduce spatial resolution. The supernet is trained for 2 epochs as in \cite{liu2020unnas}.
\begin{figure}[t]
\centering
    \begin{subfigure}[t]{0.45\linewidth}
    \centering\includegraphics[width=0.8\linewidth]{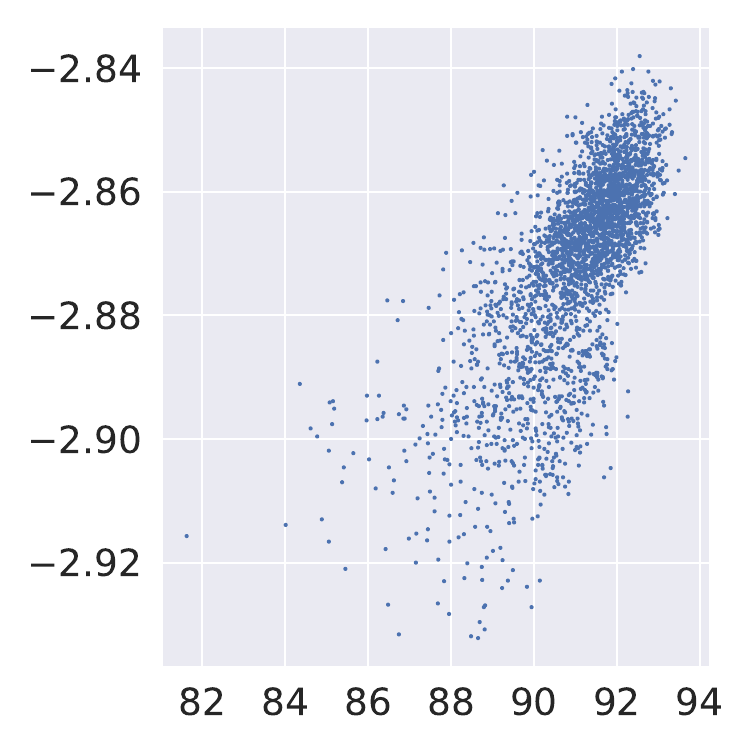}\vspace{-5pt}
    \caption{\small BossNAS on \textbf{CIFAR-10}.}\label{fig:correlation_boss_c10}
  \end{subfigure}~~~~~~~~%
  \begin{subfigure}[t]{0.45\linewidth}
    \centering\includegraphics[width=0.8\linewidth]{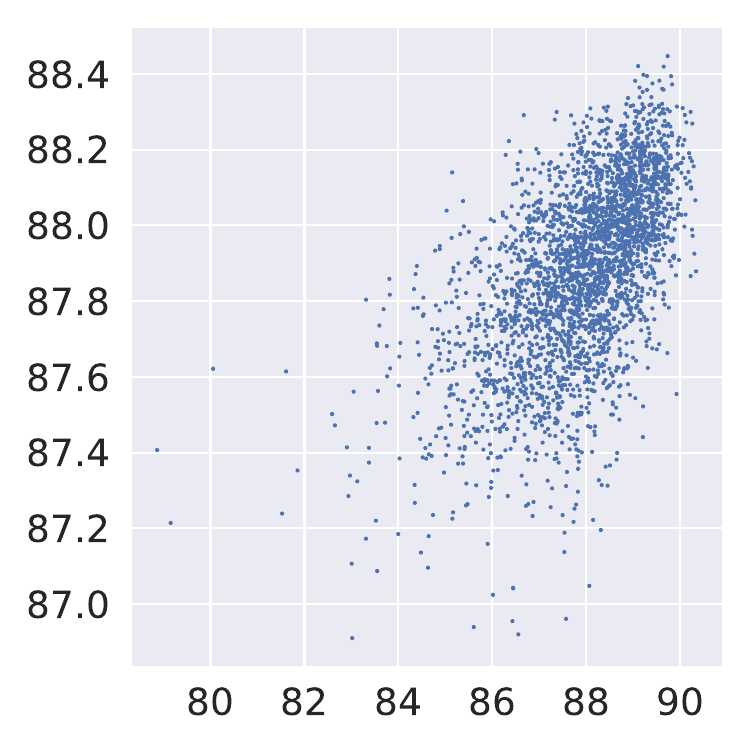}\vspace{-5pt}
    \caption{\small CE \cite{hesslow2021contrastive} on \textbf{CIFAR-10}.}\label{fig:correlation_ce_c10}
  \end{subfigure}\\
  \begin{subfigure}[t]{0.45\linewidth}
    \centering\includegraphics[width=0.8\linewidth]{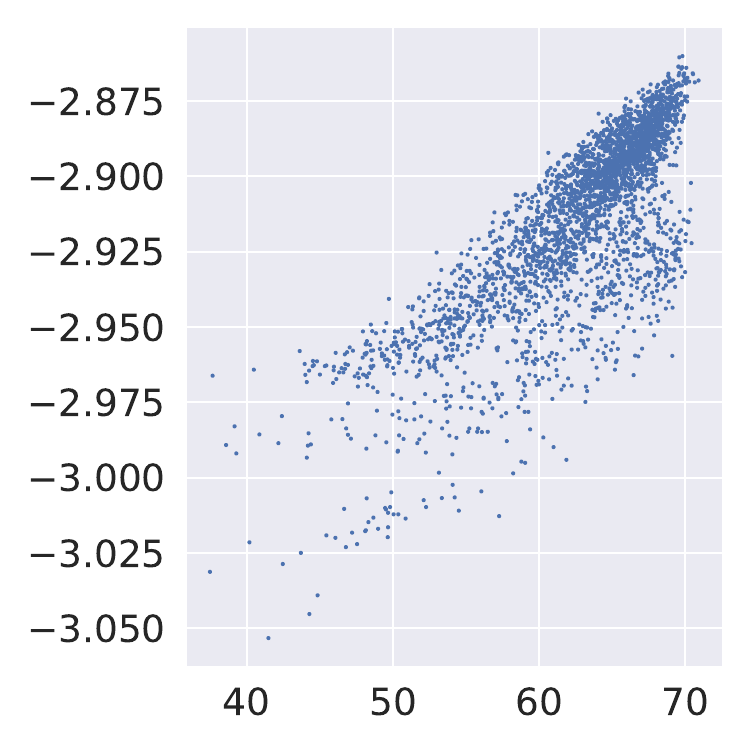}\vspace{-5pt}
    \caption{\small BossNAS on \textbf{CIFAR-100}.}\label{fig:correlation_boss_c100}
  \end{subfigure}~~~~~~~~%
  \begin{subfigure}[t]{0.45\linewidth}
    \centering\includegraphics[width=0.8\linewidth]{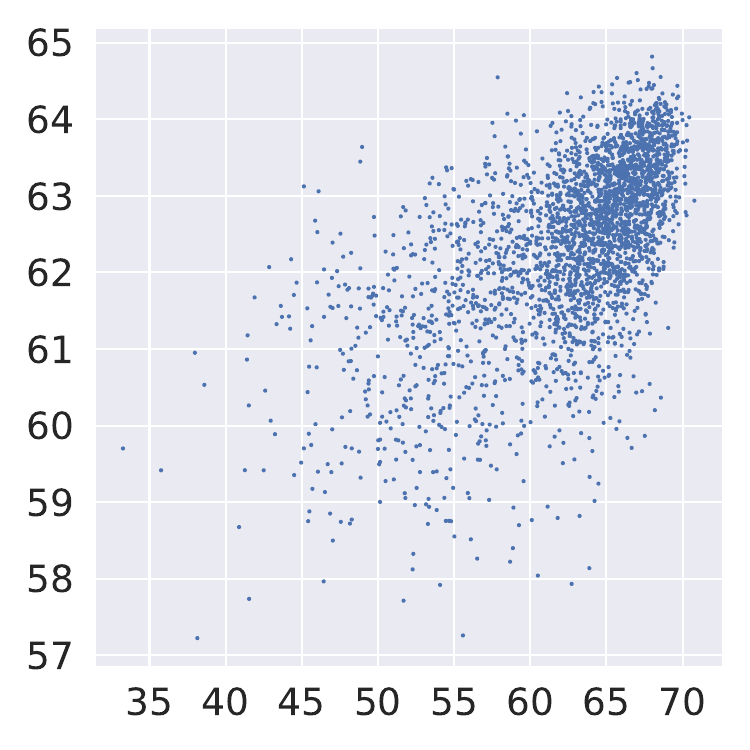}\vspace{-5pt}
    \caption{\small CE \cite{hesslow2021contrastive} on \textbf{CIFAR-100}.}\label{fig:correlation_ce_c100}
  \end{subfigure}\\
  \caption{Comparison of architecture rating and its true accuracy of our BossNAS and CE \cite{hesslow2021contrastive} on \textbf{NATS-Bench} $\boldsymbol{\mathcal{S_S}}$ with \textbf{CIFAR datasets}.}\label{fig:correlation_nats}
\end{figure}
\begin{table}[t]
    \small
    \centering
    \begin{tabular}{l|l|c|c|c}
        \toprule
        Dataset & Method      &$\normalsize\tau$         & $\normalsize\rho$        & $\normalsize R$ \\
        \midrule
         \multirow{2}{*}{CIFAR-10}& CE  \cite{hesslow2021contrastive}& 0.42           &0.60           &0.59\\
         & \textbf{BossNAS}  & \textbf{0.53}  &\textbf{0.73}  &\textbf{0.72}\\\midrule
         \multirow{2}{*}{CIFAR-100}& CE  \cite{hesslow2021contrastive}& 0.43           &0.60           &0.60\\
         & \textbf{BossNAS}  & \textbf{0.59}  &\textbf{0.76}  &\textbf{0.79}\\
        \bottomrule
    \end{tabular}
    \vspace{-7pt}
    \caption{Architecture rating accuracy on \textbf{NATS-Bench} $\boldsymbol{\mathcal{S_S}}$ with \textbf{CIFAR datasets}.
    }
    \label{tab:correlation_cifar100}
    \vspace{-15pt}
\end{table}
\subsection{Additional Analysis on NATS-Bench $\boldsymbol{\mathcal{S_S}}$}\label{sec:app_nats}
\noindent\textbf{Architecture rating comparison.}
We compare with the predictor-based NAS method CE \cite{hesslow2021contrastive} by architecture rating accuracy on CIFAR-10 and CIFAR-100. As shown in Fig. \ref{fig:correlation_nats}, we compare the two NAS methods by plotting the correlation of the architecture rating and the true accuracy of 3000 randomly sampled architectures from NATS-Bench \textit{size} search space $\mathcal{S_S}$ \cite{dong2021nats}. Architectures with BossNAS form denser and more spindly scatter pattern than CE on both of the two datasets. Moreover, as measured quantitatively in Tab. \ref{tab:correlation_cifar100}, BossNAS outperforms CE by a large margin (\textbf{0.11} and \textbf{0.16} $\normalsize\boldsymbol\tau$) in both datasets.

\begin{figure}[t]
\centering
\begin{subfigure}[t]{\linewidth}
    \centering\includegraphics[width=0.95\linewidth]{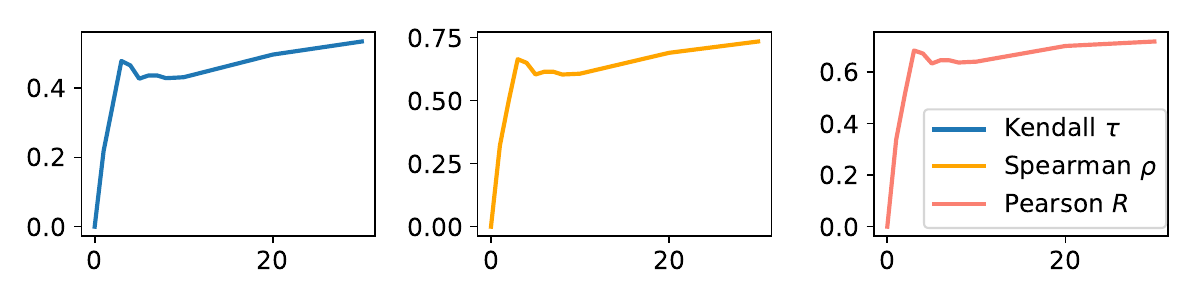}\vspace{-8pt}
    \caption{\small Ranking correlations during supernet training on \textbf{CIFAR-10}.}\label{fig:convergence_c10}
  \end{subfigure}\\\vspace{0pt}
  \begin{subfigure}[t]{\linewidth}
    \centering\includegraphics[width=0.95\linewidth]{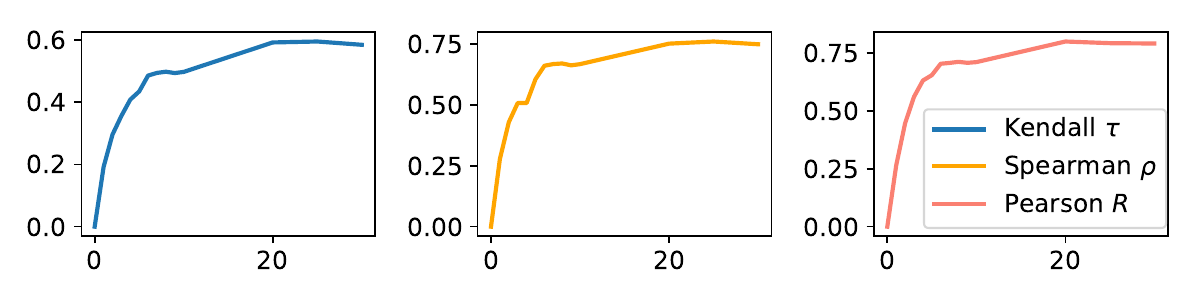}\vspace{-8pt}
    \caption{\small Ranking correlations during supernet training with \textbf{CIFAR-100}.}\label{fig:convergence_c100}
  \end{subfigure}\\
  \vspace{-5pt}
  \caption{Convergence behavior of BossNAS on \textbf{NATS-Bench} $\boldsymbol{\mathcal{S_S}}$ and \textbf{CIFAR datasets}.}\label{fig:convergence_nats}
\end{figure}

\noindent\textbf{Convergence Behavior.}
We illustrate the architecture rating accuracy of BossNAS during its 30 epoch supernet training phase on CIFAR datasets in Fig. \ref{fig:convergence_nats}. The architecture rating accuracy increases quickly and steadily with minor fluctuations, in a similar manner with that on MBConv search space (Fig. \ref{fig:convergence}). In particular, architecture rating accuracy of our BossNAS converges to a satisfactory result, \textbf{0.76} $\normalsize\boldsymbol\rho$, smoothly and quickly within only 20 epochs on CIFAR-100, and continues to be stable for the subsequent 10 epochs.
\subsection{Visualization of Human-designed Architectures in HyTra}
\begin{figure}
\centering
\begin{subfigure}[t]{\linewidth}
    \centering\includegraphics[width=0.9\linewidth]{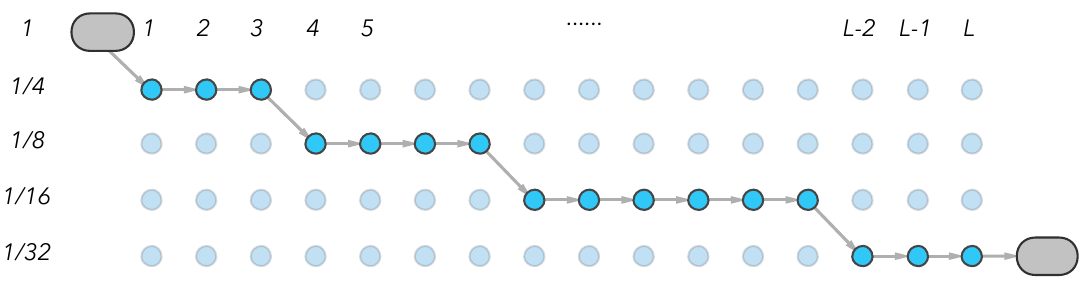}
    \caption{\small Architecture of ResNet50-T.}\label{fig:resnet-t}
  \end{subfigure}\\
  \begin{subfigure}[t]{\linewidth}
    \centering\includegraphics[width=0.9\linewidth]{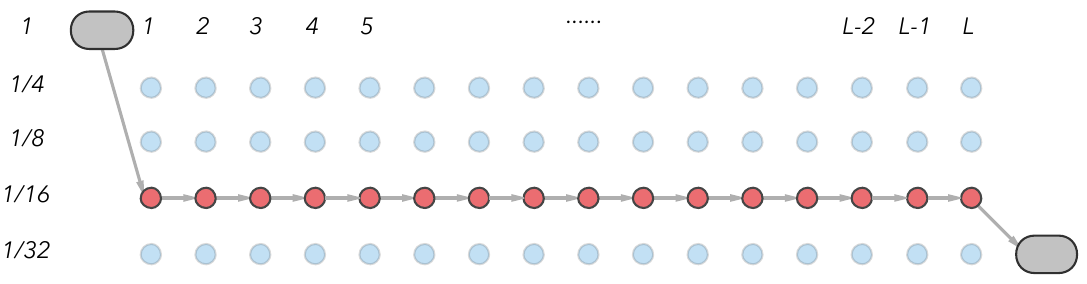}
    \caption{\small Architecture of ViT-T/16.}\label{fig:vit-t}
  \end{subfigure}\\
    \begin{subfigure}[t]{\linewidth}
    \centering\includegraphics[width=0.9\linewidth]{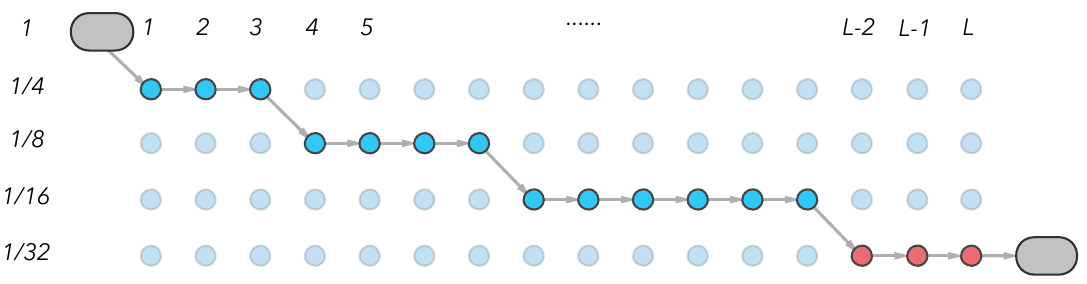}
    \caption{\small Architecture of BoTNet-T.}\label{fig:botnet-t}
  \end{subfigure}\\
  \caption{Visualization of Human-designed Architectures in HyTra. \textcolor{cyan}{Blue} nodes denotes \textcolor{cyan}{$\normalsize\mathtt{ResConv}$} and \textcolor{red}{red} nodes denotes \textcolor{red}{$\normalsize\mathtt{ResAtt}$}.}\label{fig:visualize-manual}
\vspace{-19pt}
\end{figure}

The architectures of ResNet50-T, ViT-T/16 and BoTNet50-T from our HyTra search space are illustrated in Fig. \ref{fig:visualize-manual}. Their architectures follow as closely as possible to the architectures of their prototypes.%
\end{document}